\documentclass{article}

\PassOptionsToPackage{numbers, sort}{natbib}

\usepackage[preprint]{neurips_2021}

\usepackage{algpseudocodex/algpseudocodex}

\usepackage{mathtools}
\usepackage{xspace}
\usepackage{amsmath}
\usepackage{bbm}
\usepackage{amssymb}
\usepackage{pifont}
\usepackage{caption}
\usepackage{subcaption}
\usepackage{graphicx}
\usepackage{wrapfig}
\usepackage{algorithm}
\usepackage{enumitem}

\usepackage[utf8]{inputenc} %
\usepackage[T1]{fontenc}    %
\usepackage{hyperref}       %
\usepackage{url}            %
\usepackage{booktabs}       %
\usepackage{amsfonts}       %
\usepackage{nicefrac}       %
\usepackage{microtype}      %
\usepackage{xcolor}         %

\usepackage{hyperref}

\newcommand{\ptilde}{\Tilde{p}}

\newcommand{\rhat}{\hat{r}}

\newcommand{\btheta}{\boldsymbol{\theta}}
\newcommand{\bvartheta}{\boldsymbol{\vartheta}}
\newcommand{\boldeta}{\boldsymbol{\eta}}

\newcommand{\bx}{\boldsymbol{x}}
\newcommand{\by}{\boldsymbol{y}}
\newcommand{\bz}{\boldsymbol{z}}

\newcommand{\bX}{\boldsymbol{X}}
\newcommand{\bY}{\boldsymbol{Y}}

\newcommand{\brho}{\boldsymbol{\rho}}

\newcommand{\bg}{\boldsymbol{g}}
\newcommand{\bSigma}{\boldsymbol{\Sigma}}

\newcommand{\indicator}{\mathbbm{1}}
\newcommand{\identity}{\boldsymbol{I}}

\newcommand{\cmark}{\ding{51}}
\newcommand{\xmark}{\ding{55}}
\newcommand{\greencheck}{{\color{green}\cmark}}
\newcommand{\redx}{{\color{red}\xmark}}
\newcommand{\yellowcirc}{{\color{blue}\textbullet}}

\DeclareMathOperator*{\argmax}{arg\,max}
\DeclareMathOperator{\supp}{supp}

\newcommand{\colorREJABC}[1]{\textcolor[HTML]{000000}{#1}}
\newcommand{\colorNLE}[1]{\textcolor[HTML]{000000}{#1}}
\newcommand{\colorNPE}[1]{\textcolor[HTML]{000000}{#1}}
\newcommand{\colorNRE}[1]{\textcolor[HTML]{000000}{#1}}
\newcommand{\colorSMCABC}[1]{\textcolor[HTML]{000000}{#1}}
\newcommand{\colorSNLE}[1]{\textcolor[HTML]{000000}{#1}}
\newcommand{\colorSNPE}[1]{\textcolor[HTML]{000000}{#1}}
\newcommand{\colorSNRE}[1]{\textcolor[HTML]{000000}{#1}}
\newcommand{\colorTMNRE}[1]{\textcolor[HTML]{000000}{#1}}

\newcommand{\REJABC}{\colorREJABC{\textsc{rej-abc}}}
\newcommand{\NLE}{\colorNLE{\textsc{nle}}}
\newcommand{\NPE}{\colorNPE{\textsc{npe}}}
\newcommand{\NRE}{\colorNRE{\textsc{nre}\xspace}}
\newcommand{\MNRE}{\colorTMNRE{\textsc{mnre}}}
\newcommand{\TMNRE}{\colorTMNRE{\textsc{tmnre}}}
\newcommand{\NREA}{\colorNRE{\textsc{nre\_a}}}
\newcommand{\NREB}{\colorNRE{\textsc{nre\_b}}}
\newcommand{\SMCABC}{\colorSMCABC{\textsc{smc-abc}}}
\newcommand{\SNLE}{\colorSNLE{\textsc{snle}}}
\newcommand{\SNPE}{\colorSNPE{\textsc{snpe}}}
\newcommand{\SNRE}{\colorSNRE{\textsc{snre}}}
\newcommand{\SMNRE}{\colorSNRE{\textsc{smnre}}}

\newcommand{\SNREB}{\colorNRE{\textsc{snre\_b}}}

\newcommand{\MCMC}{\textsc{mcmc}}
\newcommand{\ABC}{\textsc{abc}}
\newcommand{\SBI}{\textsc{sbi\xspace}}

\title{Truncated Marginal Neural Ratio Estimation}

\author{%
Benjamin Kurt Miller \\
University of Amsterdam \\
\texttt{b.k.miller@uva.nl}
\And
Alex Cole \\
University of Amsterdam \\
\texttt{a.e.cole@uva.nl}
\And
Patrick Forr\'e \\
University of Amsterdam \\
\texttt{p.d.forre@uva.nl} \\
\And
Gilles Louppe \\
University of Li\`ege \\
\texttt{g.louppe@uliege.be} \\
\And
Christoph Weniger \\
University of Amsterdam \\
\texttt{c.weniger@uva.nl} \\
}

\begin{document}

\maketitle

\begin{abstract}
Parametric stochastic simulators are ubiquitous in science, often featuring high-dimensional input parameters and/or an intractable likelihood. Performing Bayesian parameter inference in this context can be challenging. 
We present a neural simulation-based inference algorithm which simultaneously offers simulation efficiency and fast empirical posterior testability, which is unique among modern algorithms.
Our approach is simulation efficient by simultaneously estimating low-dimensional marginal posteriors instead of the joint posterior and by proposing simulations targeted to an observation of interest via a prior suitably truncated by an indicator function.  Furthermore, by estimating a locally amortized posterior our algorithm enables efficient empirical tests of the robustness of the inference results.  
Since scientists cannot access the ground truth, these tests are necessary for trusting inference in real-world applications.
We perform experiments on a marginalized version of the simulation-based inference benchmark and two complex and narrow posteriors, highlighting the simulator efficiency of our algorithm as well as the quality of the estimated marginal posteriors. Implementation on GitHub. \footnote{%
    Implementation of experiments at \href{https://github.com/bkmi/tmnre/}{https://github.com/bkmi/tmnre/}. Ready-to-use implementation of underlying algorithm at \href{https://github.com/undark-lab/swyft/}{https://github.com/undark-lab/swyft/}.
}
\end{abstract}

\section{Introduction}
Parametric stochastic simulators are ubiquitous in science \cite{Banik_2018, alsing2019fast, brehmer2018constraining} and using them to solve the Bayesian inverse problem is of general interest.
Likelihood-based methods like Markov chain Monte Carlo (\MCMC) \cite{metropolis,hastings} or nested sampling \cite{Skilling2006} are applicable when the likelihood is tractable.
It is equally common that the likelihood is only implicitly defined by the simulator or is inefficient to compute. For this so-called \emph{likelihood-free} or \emph{simulation-based} inference, the traditional approach is Approximate Bayesian Computation (\ABC) \cite{first_abc, second_abc}. See \cite{sisson2018handbook} for a reference.

Simulation-based inference (\SBI) is closely connected to \ABC\ and has been an open research topic since as early as the 1980s \cite{diggle1984monte}. Deep learning has accelerated progress in the field~\cite{Hermans2019, Cranmer2020, papamakarios2019sequential, greenberg2019automatic}.  
Proposed algorithms that learn the likelihood~\cite{papamakarios2019sequential} or the posterior~\cite{epsilon_free, lueckmann2017flexible, greenberg2019automatic} utilize a density estimator. The likelihood-to-evidence ratio~\cite{Hermans2019} can be learned via a classification-based technique. Refs.\ \cite{Hermans2019} and \cite{greenberg2019automatic} were brought into a unified framework by \cite{Durkan2020}.

High-fidelity simulators often have many parameters and/or an intractable likelihood function, which can make inference notoriously difficult.
Practitioners are usually faced with observational data and an expensive stochastic simulator, without access to the ground truth posterior. They want a testably accurate posterior estimate without extreme simulation expense. 
With existing methods, the practitioner must choose between increased accuracy per simulation (so-called sequential methods \cite{papamakarios2019sequential, sbibm}) or efficient empirical testability (so-called amortized methods \cite{Hermans2019}). We provide a method which offers both simultaneously with a balance that can be tuned by a hyperparameter. Three attributes contribute to this goal:
\begin{description}[font=\textit]
 	\item[Targeted inference.]
 	Focusing simulations on the parameter regions that are most relevant for the inference problem and target data is more efficient. This is particularly true when most posterior density is concentrated compared to the prior's density.
	\item[Marginal posteriors instead of the joint.]
	Scientific insight is often based on a low dimensional marginalization of the posterior with nuisance parameters removed \cite{aghanim2020planck}.
    The additional information of the full joint posterior might not be worth the additional cost afforded.
	Targeting marginals directly, by estimating only the marginal for the parameters of interest, is simpler and sufficient for many scientific, parameter estimation, and bounding purposes.
	\item[Consistency checks through local amortization.]
	Practitioners are interested in testing the quality of inference methods~\cite{feldman1998unified, geringer2015comprehensive, algeri2016method}. One such test is to compare the empirical and nominal contained mass of estimated credible regions \cite{hermans2021averting}.
	\emph{Amortized} methods learn the posterior for any data, generated by any parameter, facilitating empirical study of the nominal credible regions on fabricated data.  Still, learning an amortized posterior is excessive if only a small subset of parameters are consistent with a target observation.  
	
	We propose the concept of \emph{local amortization} to learn the posterior on said subset, combining simulator efficiency of targeted inference with the testability of amortization. Both are critical components for enabling trustworthy scientific results.
\end{description}

\paragraph{Our contribution.} We propose an algorithm that simultaneously achieves all three of the above aspects: Truncated Marginal Neural Ratio Estimation (\TMNRE). It approximates the marginal likelihood-to-evidence ratio in a sequence of rounds and shines when the joint posterior is prohibitively costly.  As a basis, we adopt likelihood-to-evidence ratio estimation proposed in~\cite{Hermans2019}, although our truncation scheme is applicable to other neural simulation-based inference methods which estimate the posterior or likelihood \cite{Durkan2020, papamakarios2019sequential}. Our iterative scheme is loosely inspired by likelihood-based nested sampling~\cite{Skilling2006, Feroz2008, Handley2015} since we generate training data drawn from a nested sequence of truncated priors in multiple rounds.  Our algorithm (a) preferentially generates simulations in relevant regions of the parameter space, (b) allows estimation of all marginals of interest simultaneously and in parallel from the same training data, and (c) yields posteriors that are locally amortized in a constrained region around the posterior, enabling empirical self-consistency test of the inference results.

\paragraph{Related work.}

\begin{table}[t]
	\caption{Comparison of \SBI\ methods, including our proposed \TMNRE, along with select properties. Properties listed are intended to showcase \TMNRE\ and do not necessarily reflect the most desirable properties in every inference setting. For example, if cost were not a inhibiting factor a tractable joint distribution may be more appealing than targeting marginals directly. Similarly, a fully amortized posterior estimate is more flexible than a targeted one but remains, often, prohibitively expensive.}
	\label{table:feature_comparison}
	\centering
	\resizebox{\textwidth}{!}{
    	\begin{tabular}{lccccccc}
    		\toprule
    		Property / Method & Likelihood-based & \ABC & \NRE & \NPE & \SNRE & \SNPE & \textbf{\TMNRE} \\
    		\midrule
    		Targeted inference &  \greencheck & \yellowcirc & \redx & \redx & \greencheck & \greencheck & \greencheck \\
    		Simulator efficient \emph{direct} marginals & \redx  & \greencheck & \yellowcirc & \yellowcirc & \redx & \redx & \greencheck \\
    		(Local) amortization & \redx & \redx & \greencheck & \greencheck & \redx & \redx & \greencheck \\
    		\bottomrule
    	\end{tabular}
	}
\end{table}

In Table~\ref{table:feature_comparison}, we compare the properties and features of a selection of deep-learning based simulation-based inference methods that are directly relevant for our work. 
Sampling from regions of highest probability density is baked into most \textit{likelihood-based} methods \cite{metropolis, hastings, Skilling2006, Handley2015, grenander1994representations, roberts1996exponential, roberts1998optimal, Feroz2008}. Amortization is generally not available with these methods because they sample from a particular posterior.
\textit{Approximate Bayesian Computation (\ABC)} is a rejection sampling technique where proposed samples from the generative model are accepted based on a user defined distance criterion comparing generated data to the observation of interest. Two important methods include \REJABC\ \cite{sisson2018handbook} where the proposal distribution is simply the parameter prior and \SMCABC\ \cite{Toni2009-fd, Beaumont2009-gl} where the proposal is iteratively refined. 
Blum and Fran\c{c}ois \cite{blum2010non} introduce an \ABC\ distance criterion weighting mechanism to tune the posterior sampler as well as a proposal prior which draws from a truncated region of true prior. It estimates the support of previously-accepted samples via support vector machines \cite{scholkopf2002learning} and samples from this region with rejection.

Likelihood-free inference can be cast as a conditional density estimation problem targeting either the posterior directly \cite{greenberg2019automatic, epsilon_free, lueckmann2017flexible, Durkan2020} or the likelihood \cite{papamakarios2019sequential, lueckmann2019likelihood}. This technique requires a density estimator, normally implemented as a mixture density network \cite{bishop} or a normalizing flow \cite{papamakarios2017masked, papamakarios2019normalizing}. \textit{Neural Likelihood Estimation} (\NLE) performs well on benchmark tasks but must learn a density representation of the data in an unsupervised setting--difficult for complex data. Modern variants of \textit{Neural Posterior Estimation} (\NPE) \cite{Durkan2020} have become effective enough to offer an alternative estimation method for scientific practitioners \cite{dax2021real}.

\textit{Amortized Approximate Ratio Estimators / Neural Ratio Estimation (\NRE):} Binary classification allows estimation of the likelihood ratio between two hypotheses \cite{Cranmer2020, Cranmer2015, Mohamed2016, gutmann2016bayesian, thomas2016likelihood, tran2017hierarchical} and was most famously applied to Generative Adversarial Networks \cite{goodfellow2014generative}. Ref.~\cite{Hermans2019} noted that naive application in the likelihood-free setting was unsatisfactory because the mathematically arbitrary choice of reference hypothesis significantly affected empirical \MCMC\ results. Comparing likelihoods from jointly drawn $(\bx,\btheta) \sim p(\bx, \btheta)$ and marginally drawn $(\bx,\btheta) \sim p(\bx)p( \btheta)$ samples, where $\bx$ and $\btheta$ refer respectively to simulated data and simulator parameters, addresses the issue.
Ref. \cite{Durkan2020} cast \NRE\ and \NPE\ in a unifying framework by adapting the loss function to contrast several possible hypothetical parameters.
In this paper we refer to the algorithm described in \cite{Hermans2019} as \NRE\ or \NREA\ while the likelihood ratio algorithm described in \cite{Durkan2020} is referred to as \NREB.

\textit{Directly estimating the marginal posterior} distribution has been mentioned \cite{Hermans2019} and explored \cite{swyft, moment_networks}.
Moment networks \cite{moment_networks} produce the (central) moments of the posterior distribution, without calculating the density explicitly, via a hierarchy of neural networks trained on a regression problem. Ref. \cite{moment_networks} also introduced a method which learns marginal posteriors with normalizing flows but does not address targeted inference or testability of estimated posteriors.

\textit{Sequential Methods:} The neural likelihood-free methods generally offer a so-called sequential formulation that targets the posterior of a particular observation $\bx_{o}$ \cite{Durkan2020, Hermans2019, papamakarios2019sequential}. Rather than drawing samples from the prior, the simulation budget is divided between rounds and the previous round's posterior is used as the new proposal distribution for the next round. This method increases simulation efficiency, but does not allow for amortization.  Importantly, sequential methods can become highly inefficient when targeting multiple marginal posteriors directly because the previous round's marginal posterior does not update beliefs about the other parameters. A full parameter vector is necessary to run the simulator, thus defeating the purpose for all nuisance (marginalized-over) parameters.

\section{Method}
\label{sec:method}
We aim to estimate any marginal posterior of interest using an approximate marginal likelihood-to-evidence ratio. Although we normally compute every one- and two-dimensional marginal posterior for visualization purposes, our method is not limited to this restriction. We define the object of study...

Let parametric stochastic simulator $\bg$ be a nonlinear function that maps a vector of real parameters $\btheta = (\theta_{1}, \dots, \theta_{D})$ and a stochastic latent state $\bz \in \mathbb{R}^{N_z}, N_z \in \mathbb{N}$ to an observation $\bx = \bg(\btheta, \bz)$. 
We consider a \emph{factorizable prior} $p(\btheta) = p(\theta_{1}) \cdots p(\theta_{D})$ over the parameters. The joint posterior is given via Bayes' rule as $p(\btheta \mid \bx) = p(\bx \mid \btheta)p(\btheta)/p(\bx) $, where $p(\bx \mid \btheta)$ is the intractable likelihood (or implicit distribution~\cite{diggle1984monte, hartig2011statistical}) and $p(\bx)$ is the evidence.
Our goal is to efficiently compute arbitrary marginal posteriors, $p(\bvartheta \mid \bx)$.  
Here, $\bvartheta$ are the parameters of interest, and we denote all other (nuisance) parameters by $\boldeta$, such that $\btheta = (\bvartheta, \boldeta)$.  The marginal posterior is obtained from the joint distribution $p(\bvartheta, \boldeta \mid \bx) \coloneqq p(\btheta \mid \bx)$ by integrating over all components of $\boldeta$,
\begin{equation}
    p(\bvartheta \mid \bx) 
    = \int p(\bvartheta, \boldeta \mid \bx) \, d\boldeta
    = \frac{\int p(\bx \mid \bvartheta, \boldeta) p(\boldeta) \, d\boldeta}{p(\bx)} p(\bvartheta)
    \coloneqq \frac{p(\bx \mid \bvartheta)}{p(\bx)} p(\bvartheta).
\label{eqn:posterior}
\end{equation}
where we used Bayes' rule, prior factorizibility, and defined the marginal likelihood $p(\bx \mid \bvartheta)$.

\subsection{Marginal Neural Ratio Estimation (\MNRE)}
\label{sec:mnre}

This paper considers the set of one- and two-dimensional marginal posteriors and their corresponding parameters of interest. Given parameter vector $\btheta \in \mathbb{R}^{D}$, define the set of all parameters associated with the one dimensional marginal posteriors by $\Theta_{1} \coloneqq \left\{ \theta_{1}, \ldots, \theta_{D} \right\}$. 
We do something similar, up to symmetry, for all two dimensional marginal posteriors $\Theta_{2} \coloneqq \left\{ (\theta_{i}, \theta_{j}) \in \mathbb{R}^{2} \; \middle| \; i = 1, \ldots, D, \, j =i + 1, \ldots, D \right\}$. 
We set our marginals of interest $\{\bvartheta_{k}\} \coloneqq \Theta_{1} \cup \Theta_{2}$ but in the general case, $\{\bvartheta_{k}\}$ can be any set of marginals that the practitioner desires. For every $\bvartheta_{k}$ we use \NRE\ \cite{Hermans2019} to estimate the corresponding marginal likelihood-to-evidence ratio
\begin{equation}
    r_k(\bx \mid \bvartheta_k) \coloneqq \frac{p(\bx \mid \bvartheta_k)}{p(\bx)}
    =\frac{p(\bx,\bvartheta_k)}{p(\bx)p(\bvartheta_k)}
    =\frac{p(\bvartheta_k \mid \bx)}{p(\bvartheta_k)}
    \;.
\end{equation}
To this end, we train binary classifiers $\hat{\rho}_{k,\phi}(\bx, \bvartheta_k)$ to distinguish jointly drawn parameter-simulation pairs $(\bx, \bvartheta_k) \sim p(\bx, \bvartheta_k)$ from marginally drawn parameter-simulation pairs $(\bx, \bvartheta_k) \sim p(\bx) p(\bvartheta_k)$, where $\phi$ represents the parameters of the classifier.
A Bayes optimal classifier $\rho_{k}$ would recover the density $\rho_{k}(\bx, \bvartheta_k) = \frac{p(\bx, \bvartheta_k)}{p(\bx, \bvartheta_k) + p(\bx) p(\bvartheta_k)}$. 
Then the ratios of interest can be estimated by
\begin{equation}
\label{eqn:define_ratio_estimator}
	\hat r_k(\bx \mid \bvartheta_k) \coloneqq
	\frac{\hat{\rho}_{k,\phi}(\bx, \bvartheta_k)}{1 - \hat{\rho}_{k,\phi}(\bx, \bvartheta_k)} 
	\approx \frac{p(\bx, \bvartheta_k)}{p(\bx) p(\bvartheta_k)} 
	= r_k(\bx \mid \bvartheta_k)\;.
\end{equation}
We train each ratio estimator $\hat{r}_k(\bx \mid \bvartheta)$ using Adam \cite{kingma2015adam} to minimize the binary cross-entropy (\textsc{bce})
\begin{equation}
\label{eqn:loss}
    \ell_{k} 
    = -\int \left[ p(\bx \mid \btheta) p(\btheta) \ln \hat{\rho}_{k,\phi}(\bx, \bvartheta_k) + p(\bx)p(\btheta) \ln \left(1-\hat{\rho}_{k,\phi}(\bx,\bvartheta_k\right)) \right] d\bx\, d\btheta\;.
\end{equation}
In practice, we concatenate $\bx$ with $\bvartheta_k$ as the input to $\hat{\rho}_{k,\phi}$. Since each classifier trains independently, it is trivial to train them all in parallel using the same underlying $(\bx, \btheta)$ pairs. 

Practically, we parameterize the classifier by $\hat{\rho}_{k,\phi}(\bx, \bvartheta_k) = \sigma \circ f_{k,\phi}(\bx, \bvartheta_k)$, where $\sigma$ is the logistic sigmoid and $f_{k,\phi}$ is a neural network. The connection in Eq.~\eqref{eqn:define_ratio_estimator} between the estimated ratio and the classifier implies that $\log \hat r_k(\bx \mid \bvartheta_k) = f_{k,\phi}(\bx, \bvartheta_k)$. We call the above technique \MNRE.

When training data is limited, we found empirically (see Sec.~\ref{sec:torus} below) that the \MNRE\ approach typically leads to conservative (i.e., not overconfident / not narrower than the ground truth) likelihood-to-evidence ratio estimates, provided early stopping criteria are used to avoid over-fitting of the classifier.
At its core \MNRE\ solves a simple, supervised binary classification task rather than a complex, unsupervised density estimation problem.  Classification tasks are generally easier to train~\cite{Cranmer2020}, and can rely on battle-tested network architectures.

\subsection{Truncated Marginal Neural Ratio Estimation (\TMNRE)}
\label{sec:tmnre}

\MNRE\ and \NRE\ estimate a (marginal) likelihood-to-evidence ratio agnostic to the observed data $\bx$ or parameter $\btheta$, a so-called \emph{amortized} estimate. In other words, \MNRE\ is suitable when $\bx \in \left\{ \bg(\btheta, \bz) \; \middle| \; \btheta \in \Omega, \bz \in \mathbb{R}^{N_{z}} \right\}$ where $\Omega$ is the support of the prior.
We propose an extension of this algorithm that enables targeted simulation of parameters relevant to a given target observation $\bx_o$, and locally amortizes posteriors such that it enables empirical tests of the inference results.
\emph{Local amortization} implies that our proposed method is suitable when $\bx \in \left\{ \bg(\btheta, \bz) \; \middle| \; \btheta \in \Gamma , \bz \in \mathbb{R}^{N_{z}} \right\}$ where the parameter region $\Gamma \subset \Omega$ is a function of $\bx_o$ and will be defined below.

We observe that values of $\btheta$ which could not have plausibly generated $\bx_{o}$ evaluate to negligible posterior density, i.e. $p(\btheta \mid \bx_o) \approx 0$, which suggests that
the corresponding parameters $\btheta$ do not significantly contribute to the marginalization in Eq.~\eqref{eqn:posterior}. 
We denote a prior that is suitably constrained to parameters with non-negligible posterior density $p(\btheta \mid \bx_o)$ by
\begin{equation}
    \label{eqn:truncated_prior}
    p_\Gamma(\btheta) \coloneqq  V^{-1} \indicator_{\Gamma}(\btheta) p(\btheta)\;,
\end{equation}
where $\indicator_{\Gamma}(\btheta)$ is an indicator function that is unity on $\Gamma \subset \Omega$ and zero otherwise, and $V^{-1}$ is a normalizing constant (which can be interpreted as the fractional volume of the truncated prior).
The subscript $\square_\Gamma$ denotes that arbitrary symbol $\square$ is based on a prior truncated by indicator function $\indicator_{\Gamma}$.

We define a rectangular indicator function $\indicator_{\Gamma^\text{rec}}$ by discarding parameters that lie in the far tails of the one dimensional marginal posteriors of our target observation $\bx_o$,
using a thresholding $\epsilon \ll 1$, via
\begin{equation}
\label{eqn:GammaRec}
    \Gamma^\text{rec} = 
    \left\{ \btheta \in \Omega  
	\; \middle| \;
	\forall d = 1, \ldots, D: \frac{p(\theta_d \mid \bx_o)}{\max_{\theta_d} 
	p(\theta_d \mid \bx_o)} > \epsilon \right\}\;.
\end{equation}
For Gaussian joint posteriors, this scheme leads to one dimensional marginal posteriors $p_\Gamma(\theta_{d} \mid \bx_o)$ that are truncated at their approximately $\pm \sqrt{-2\ln\epsilon} \, \sigma$ tail.
In general, truncation will lead to an approximation error that can be estimated as $p_{\Gamma^\text{rec}}(\btheta \mid \bx_o) = p(\btheta \mid \bx_o) + \mathcal{O}(\epsilon) \max_{\btheta} p(\btheta \mid \bx_o)$, see Appendix~\ref{apndx:derivations}. 
Throughout this paper we use $\epsilon=10^{-6}$, which corresponds to $\pm5.26\sigma$ for a Gaussian posterior.
Those truncations do not affect the location of high-probability credible contours and have hence no practical effect on parameter inference tasks.
We provide more exemplary error estimates for a range of cases in Appendix~\ref{apndx:derivations}.

\begin{algorithm}[thb]
	\caption{Truncated Marginal Neural Ratio Estimation (TMNRE)}
	\label{alg:tmnre}
	\begin{tabular}{rp{0.85\columnwidth}}
		\textit{Inputs:}&
			Simulator $p(\bx \mid \btheta)$,
			factorizable prior $p(\btheta)$, 
			real observation $\bx_{0}$, 
			max rounds $M$,
			training data per round $N^{(m)}$,
			threshold $\epsilon$,
			dimension of parameters $D$, 
			mass ratio $\beta$,
			classifiers $\brho_{1}(\bx, \btheta) = \{ \sigma \circ f_{\phi, d}(\bx, \theta_{d})\}_{d=1}^{D}$ and $\brho_{2}(\bx, \btheta) = \{ \sigma \circ f_{\phi, d}(\bx, \bvartheta_{d}) \}_{d=(1,1)}^{(D,D)}$.
			\\
		\textit{Outputs:}& 
			Parameterized classifiers $\brho_{1}(\bx, \btheta)$ and $\brho_{2}(\bx, \btheta)$, 
			constrained region $\Gamma^\text{rec}$. \\ 
	\end{tabular}

	\begin{algorithmic}[1]
		\Procedure{mnre}{$\mathcal{D}$, $\btheta'$, $\brho_{\phi}$}
		\While{$\brho_{\phi}$ not converged}
			\State $\phi 
				\leftarrow 
				\textsc{optimizer}\left(
					\phi, 
					\nabla_\phi \sum_{k} \left[ \textsc{bce}(\hat{\rho}_{\phi, k}(\bx, \bvartheta_{k}), 1) + \textsc{bce}(\hat{\rho}_{\phi, k}(\bx, \bvartheta_{k}'), 0) \right]\right)$
		\EndWhile
		\State \Return $\mathbf{f}_{\phi}$
		\EndProcedure
	\end{algorithmic}

	\begin{tabular}{rp{0.85\columnwidth}}
	\textit{Initialize:}&
		$\mathcal{D}^{(0)} \leftarrow \{\}$, \;
		$\Gamma^{(0)} \leftarrow \supp(p(\btheta))$, \;
		$\alpha^{(0)} \leftarrow 0$, \;
		$m \leftarrow 1$.
	\end{tabular}

	\begin{algorithmic}[1]
		\Procedure{tmnre}{}
		\While{$\alpha^{(m-1)} \leq \beta$ and $m \leq M$}
			\State $\mathcal{D}_{\Gamma}^{(m-1)} \leftarrow \left\{ (\bx^{(n)}, \btheta^{(n)}) \in \mathcal{D}^{(m-1)} \; \middle| \; \btheta^{(n)} \in \Gamma^{(m-1)} \right\}$
			\Comment{Retain data in region}
			
			\State $N_{\textrm{simulate}}^{(m)} \leftarrow N^{(m)} - \lvert \mathcal{D}_{\Gamma}^{(m-1)} \rvert$ 
			\Comment{Calculate num. necessary simulations}
			
			\State $\btheta \leftarrow \{ \btheta^{(n)} \sim \indicator_{\Gamma^{(m-1)}}(\btheta) p(\btheta) \}_{n=1}^{N_{\textrm{simulate}}^{(m)}}$ 
			\Comment{Sample for jointly distributed pairs}
			
			\State $\bx \leftarrow \{ \bx^{(n)} \sim p(\bx \mid \btheta^{(n)}) \}_{n=1}^{N_{\textrm{simulate}}^{(m)}}$ 
			\Comment{Simulate jointly distributed pairs}
			
			\State $\btheta' \leftarrow \{ \btheta^{(n)} \sim \indicator_{\Gamma^{(m-1)}}(\btheta) p(\btheta) \}_{n=1}^{N^{(m)}}$ 
			\Comment{Sample for marginally distributed pairs}
			
			\State $\mathcal{D}^{(m)} \leftarrow \mathcal{D}_{\Gamma}^{(m-1)} \cup \{ (\bx^{(n)}, \btheta^{(n)}) \}_{n=1}^{N_{\textrm{simulate}}^{(m)}}$ 
			\Comment{Aggregate training data}
			
			\State $\brho_{1} \leftarrow$ \Call{mnre}{$\mathcal{D}^{(m)}$, $\btheta'$, $\brho_{1}$}
			
			\State $\Gamma^{(m)} \leftarrow \left\{ \btheta \in \Gamma^{(m - 1)}  \; \middle| \; \forall d: \frac{\hat p_{d, \Gamma^{(m)}}(\theta_d \mid \bx_o)}{\max_{\theta_d} \hat p_{d, \Gamma^{(m)}}(\theta_d \mid \bx_o)} > \epsilon \right\}$
			\Comment{Find constrained region}
			
            \State $\alpha^{(m)} \leftarrow \int \indicator_{\Gamma^{(m)}}(\btheta) p(\btheta) d\btheta / \int \indicator_{\Gamma^{(m-1)}}(\btheta) p(\btheta) d\btheta$
			\Comment{Update prior mass ratio}

			\State $m \leftarrow m + 1$
			\Comment{Increment counter}
		\EndWhile
		\State $\brho_{2} \leftarrow$ \Call{mnre}{$\mathcal{D}^{(m)}$, $\btheta'$, $\brho_{2}$}
		\State \Return $\brho_{1}$, $\brho_{2}$, $\Gamma^{(m)}$
		\EndProcedure
	\end{algorithmic}
\end{algorithm}

Our algorithm defines a series of nested indicator functions $\indicator_{\Gamma^{(m)}}$ whose regions have the property
\begin{equation}
\label{eqn:Gammas}
    \Omega \coloneqq \Gamma^{(1)}  \supset 
    \Gamma^{(2)} \supset \dots \supset \Gamma^{(M)} \supset \Gamma^\text{rec}.
\end{equation}
They iteratively approximate the indicator function $\indicator_{\Gamma^\text{rec}}$ in multiple rounds $m=1, \dots, M$.  This sequence is generated with the following steps:
\begin{itemize}
    \item We initialize $\Gamma^{(1)} = \Omega \coloneqq \supp(p(\btheta))$, meaning that we start with the unconstrained prior.  
    \item Each round $1 \leq m \leq M$, we train $D$, one dimensional ratio estimators ${\hat r_{d, \Gamma^{(m)}}(\bx \mid \theta_{d})}$ using data from within the constrained region, $\btheta \in \Gamma^{(m)}$.
    The estimated marginal posterior is $\hat p_{\Gamma^{(m)}} (\theta_{d} \mid \bx) = 
    \hat r_{d, \Gamma^{(m)}}(\bx \mid \theta_{d}) p_{\Gamma^{(m)}} (\theta_{d})$.
    To this end, do \MNRE, setting $\bvartheta_{k} = \theta_{k}, \; d \in \{1, 2, \ldots, D\}$ using the constrained prior $p_{\Gamma^{(m)}}(\btheta)$ with $N^{(m)}$ training samples per round.
    \item For each round $m<M$, we estimate the indicator function for the next round using the approximated posteriors, via
    \begin{equation}
        \label{eqn:GammaDef}
        \Gamma^{(m+1)} = 
        \left\{ \btheta \in \Gamma^{(m)}  
    	\; \middle| \;
    	\forall d: \frac{\hat p_{\Gamma^{(m)}}(\theta_d \mid \bx_o)}{\max_{\theta_d} 
    	\hat p_{\Gamma^{(m)}}(\theta_d \mid \bx_o)} > \epsilon \right\}\;.
    \end{equation}
    
    \item The last round is determined either when $m=M$ or when a stopping criterion is reached.
    The stopping criterion is defined by the ratio of consecutive truncated prior masses. It is satisfied when the sequence of truncated priors have the property $\int \indicator_{\Gamma^{(m)}}(\btheta) p(\btheta) d\btheta / \int \indicator_{\Gamma^{(m-1)}}(\btheta) p(\btheta) d\btheta > \beta$. We often set $\beta = 0.8$.
    
    \item Using the data from this final constrained region, we can approximate any marginal posterior of interest. In this paper, we estimate the one- and two-dimensional marginals necessary for corner plots. We emphasize that the data already generated during the truncation phase can be reused to learn arbitrary marginals of interest. When higher accuracy likelihood-to-evidence ratio estimates are needed, the user can simulate from the truncated region.
\end{itemize}

We briefly address failure modes. First, this algorithm relies on the assumption that posterior estimates $\hat p_{\Gamma^{(m)}}(\theta_d \mid \bx_o)$ from \MNRE\ provide a good approximation of $p(\theta_d \mid \bx_o)$.
An over-confident estimate would remove parameter ranges that are part of $\Gamma^\text{rec}$.
In practice, we have not observed this effect, although it is a concern with all simulation-based methods \cite{hermans2021averting}.  We give credit to early stopping and a conservative choice of $\epsilon$, and provide further illustration and support in Sec.~\ref{sec:experiments} below.
Second, since the truncated posterior only agrees with the ground truth up to corrections of order $\epsilon$, the iterative scheme will not converge to Eq.~\eqref{eqn:GammaRec}; rather to a similar expression where the right-hand side of the inequality in Eq.~\eqref{eqn:GammaRec} receives additional $\mathcal{O}(\epsilon)$ corrections. Although these corrections mildly affect the truncations, they are of little practical relevance since we choose an $\epsilon$ which is very small.
Both failure modes are diagnosed by checking whether high probability regions of the estimated posteriors intersect with the boundaries of the indicator function.

Algorithm~\ref{alg:tmnre} estimates the necessary marginal posteriors for corner plot visualization. We demonstrate the cost-effectiveness of this algorithm in Section~\ref{sec:experiments}. An important limitation of Algorithm~\ref{alg:tmnre} is the inaccessibility of the posterior predictive distribution. This limitation is mitigated by training a ratio estimator on all parameters within the truncated region; however, producing an accurate joint estimate may come with (sometimes significant) additional simulation costs.

Like sequential methods~\cite{papamakarios2019sequential, Durkan2020} the number of rounds $M$, the training data per round $N^{(m)}$, and any stopping criteria $\beta$ are hyperparameters. For further discussion and default values see Appendix~\ref{apndx:experiments}, for bound derivations and limitations see Appendix~\ref{apndx:derivations} and~\ref{apndx:limitations}. We present \TMNRE\ in Algorithm~\ref{alg:tmnre}.

\paragraph{Properties of our algorithm.}
We discuss the properties of our algorithm in support of Table~\ref{table:feature_comparison}.  First, our algorithm performs \textit{targeted inference} by successively focusing on regions of the parameter space that are compatible with an observation $\bx_o$.
Second, since training data is always drawn from the prior, it is possible to efficiently \textit{train arbitrary marginal posteriors} with the same training data generated for round $M$.
Third, the algorithm trains \textit{locally amortized posteriors} that are valid for parameters $\btheta \in \Gamma^{(m)}$, facilitating empirical consistency checks of the estimated posteriors within this region.
\TMNRE's properties provide a favorable cost-benefit ratio--yielding marginal insight into the posterior without paying the price of accessing the joint posterior. The price of the joint is often inhibiting for expensive simulators.
These aspects will be demonstrated in the experiments in the following section.

\section{Experiments}
\label{sec:experiments}

First, we perform experiments to compare \TMNRE\ to other algorithms on standard benchmarks from the simulation-based inference literature. Next, we highlight useful aspects of our algorithm regarding targeted inference, marginalization and local amortization with two additional experiments. These experiments compare algorithms using performance metrics which access the ground truth posterior. For practitioners who normally cannot access the ground truth, \TMNRE\ offers an empirical consistency check (see Section~\ref{sec:empirical_tests}) to assess the quality of the estimated posterior. Such a check is impractical for sequential methods and is one of the primary practical applications of \TMNRE.
Further experiments, including application on a cosmology simulator, can be found in Appendix~\ref{apndx:cosmology}.

\subsection{Performance on standard tasks}
\label{sec:sbibm}

We compare the performance of our algorithm with other traditional and neural simulation-based inference methods on a selection of problems from the \SBI\ benchmark \cite{sbibm}. Each \emph{task} is defined by a simulator, ten observations, a simulation budget, and 10,000 samples drawn from corresponding reference and approximate posterior distributions. The reference samples enable quantification of algorithmic accuracy on a range of performance metrics. We evaluate performance on all tasks except the Bernoulli Generalized Linear Model because its prior is not factorizable. Details in Appendix~\ref{apndx:experiments}.

Since our method estimates every one- and two-dimensional marginal posterior, we compare samples from our approximate marginal posteriors with samples from the reference joint posterior, marginalized over nuisance parameters. We quantify the results using the Classifier 2-Sample Test \cite{friedman2003multivariate, lopez2017revisiting}. We train a C2ST classifier for each of the $\binom{D}{1} + \binom{D}{2}$ possible one- and two-dimensional marginals, and report the mean values, and 95\% confidence intervals, in Figure~\ref{fig:sbibm}. We call this averaged performance metric \emph{C2ST-ddm}, see Appendix~\ref{apndx:evaluation_metrics} for more detail.
The results are presented as grouped by dimensionality since difficulty increases with dimension and is reflected in the C2ST-ddm scores.

\begin{wrapfigure}[38]{r}{0.6\textwidth}
	\begin{center}
    	\includegraphics[width=0.6\textwidth]{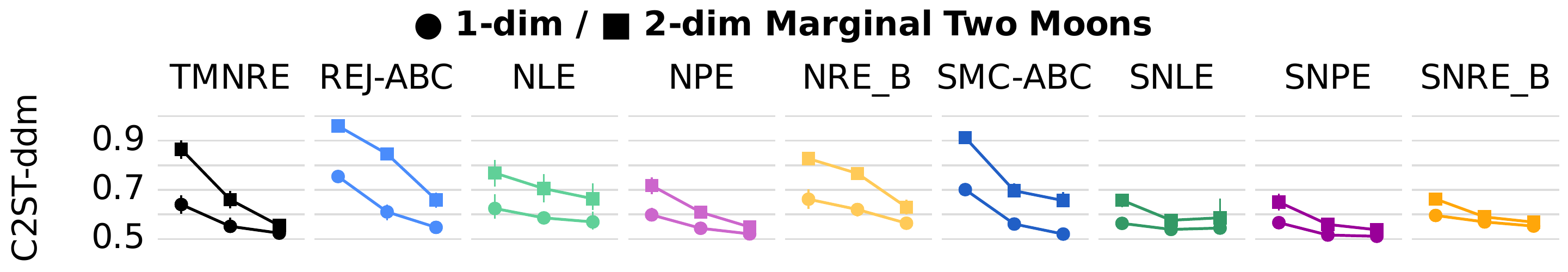}
    	\includegraphics[width=0.6\textwidth]{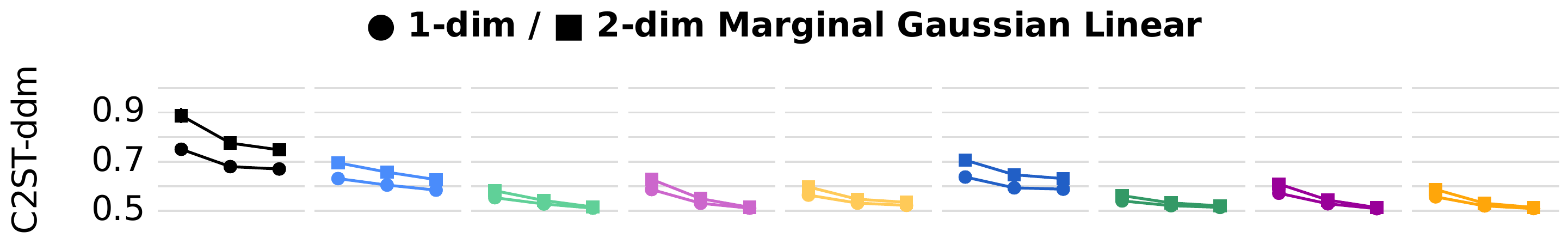}
    	\includegraphics[width=0.6\textwidth]{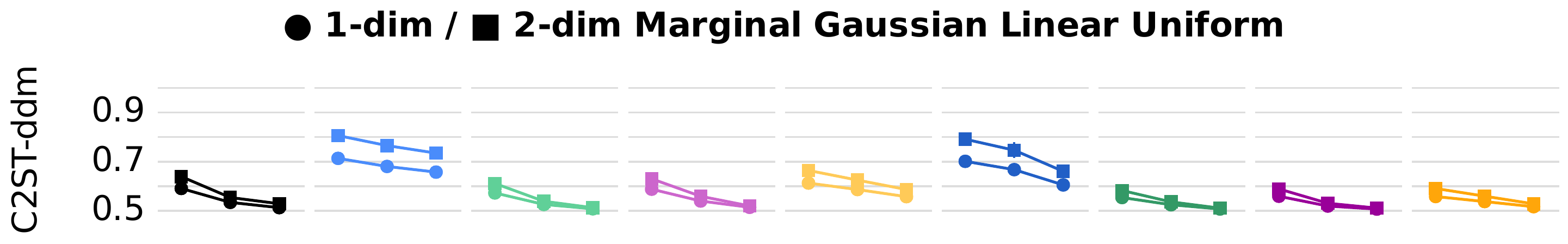}
    	\includegraphics[width=0.6\textwidth]{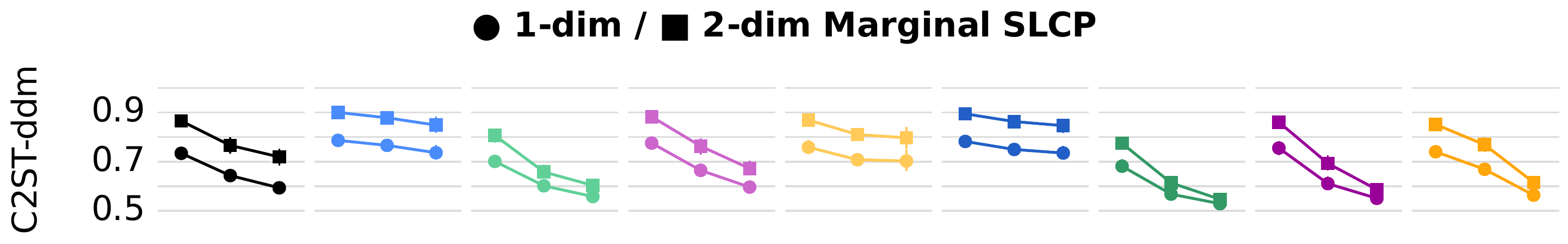}
    	\includegraphics[width=0.6\textwidth]{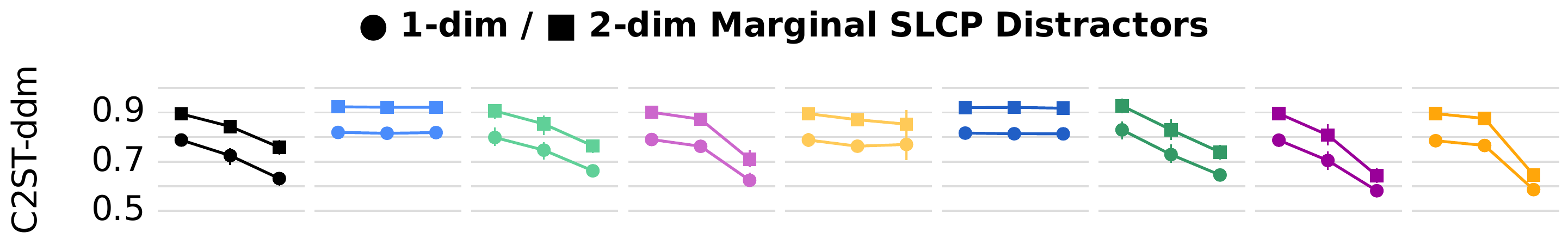}
    	\includegraphics[width=0.6\textwidth]{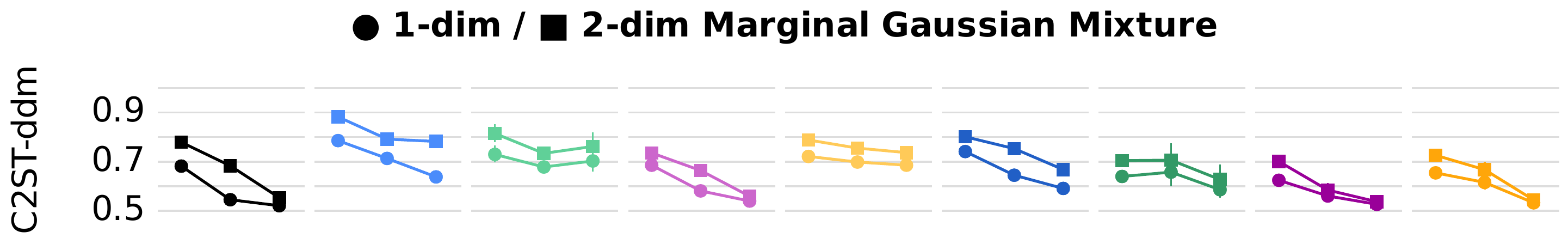}
    	\includegraphics[width=0.6\textwidth]{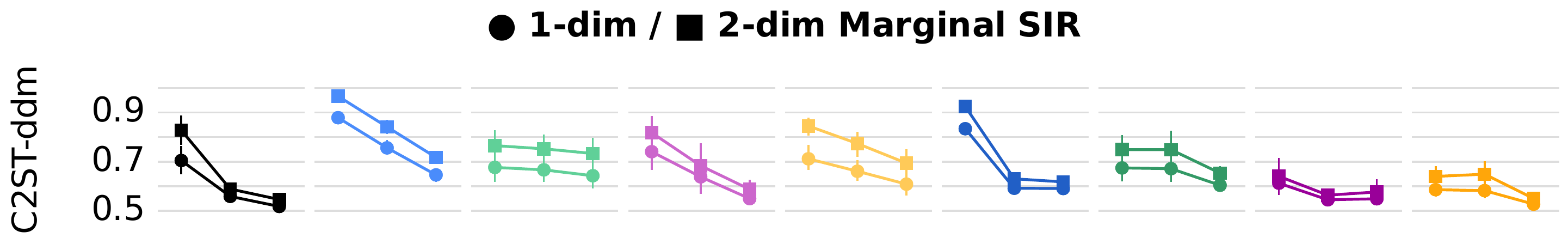}
    	\includegraphics[width=0.6\textwidth]{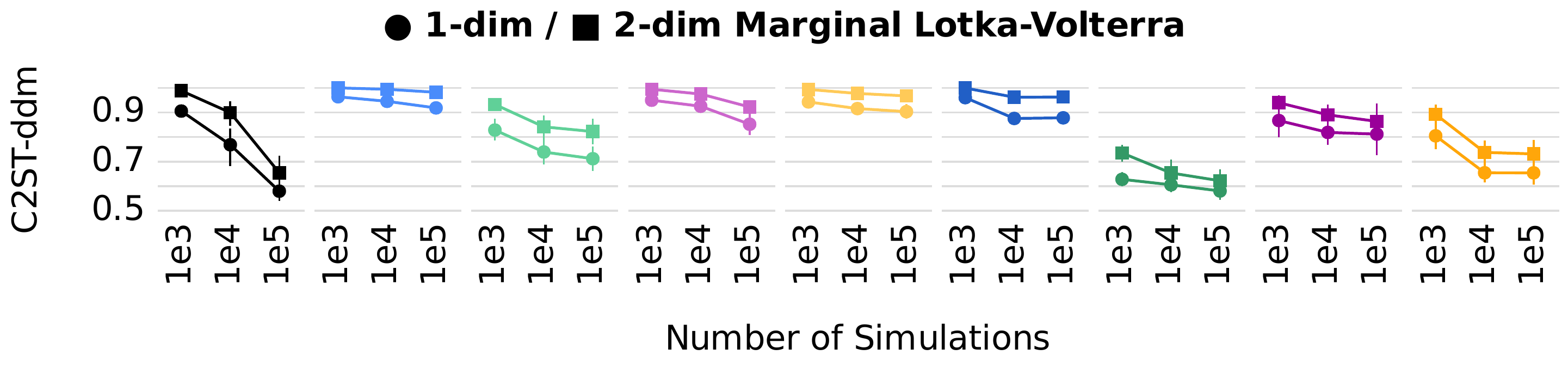}
	\end{center}
    \caption{
        \textbf{Performance on marginalized posterior benchmark tasks.} Mean and 95\% confidence intervals of classification accuracy (C2ST-ddm) for our method, \TMNRE, and other \SBI\ methods with 10 observations / budget. one- and two-dimensional scores are plotted. Lower scores imply better posteriors. Our simulation budget is approximate, see Appendix \ref{apndx:experiments}. The plot and tasks are derivative of \protect\cite{sbibm}.
	}
	\label{fig:sbibm}
\end{wrapfigure}

For comparison, we computed the C2ST-ddm on the other benchmark methods' posterior samples. Unlike our method, which was trained on the marginals directly, the benchmark methods were trained to estimate the joint posterior. We note that since \TMNRE\ trains a neural network for every marginal (efficiently, in parallel), our method has many times more parameters than any neural likelihood-free inference method that directly targets the joint. However, parameter count is not a scarce resource in \SBI. Training hyperparameters can be found in Appendix~\ref{apndx:experiments}.

The maximum number of rounds before meeting the stopping criteria varied across tasks from just one round with no truncation up to seven rounds on Gaussian Linear. Out of 240 runs, five did not converge for \TMNRE.

As shown in Figure~\ref{fig:sbibm}, our method outperformed \REJABC\ and \SMCABC\ and offers increased efficiency compared to non-sequential methods on all tasks, except Gaussian Linear. On some tasks, \TMNRE\ is competitive with sequential methods. Generally, \TMNRE\ performs best on narrow posteriors and a large simulation budget. Benefits diminished on tasks with wide posteriors like Gaussian Linear, SLCP, and SLCP Distractors--a limitation of \TMNRE. Based on these results, ours is the only method, among sequential and non-sequential techniques, which offers both sufficient accuracy and local amortization.

\subsection{Efficient targeted inference: a 3-dim torus model}
\label{sec:torus}

\begin{figure}[ht]
    \centering
    \begin{minipage}{0.25\textwidth}
        \centering
        \includegraphics[width=\textwidth]{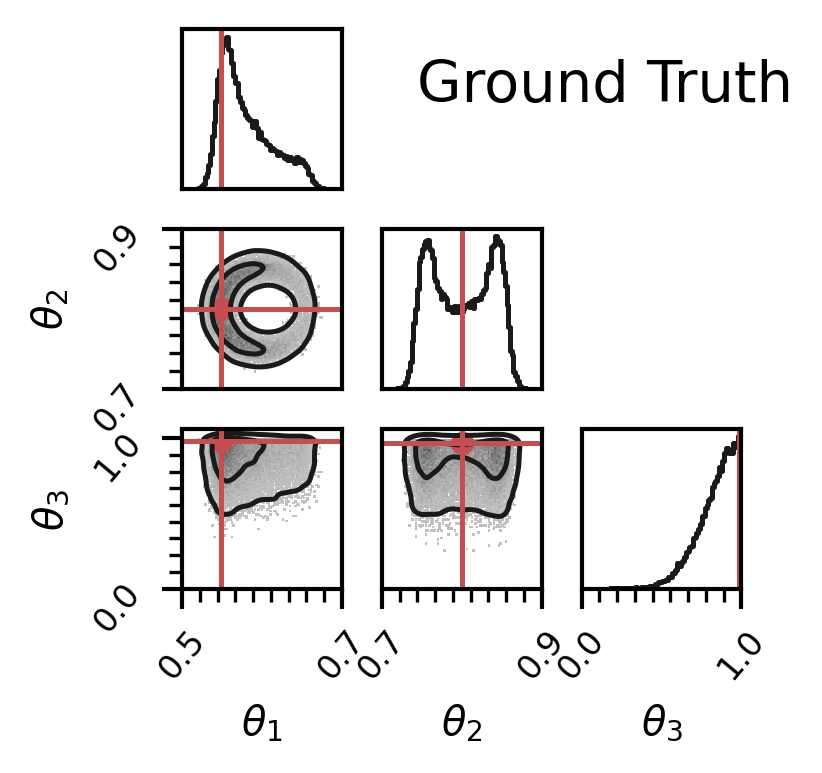}
    \end{minipage}\hfill
    \begin{minipage}{0.25\textwidth}
        \centering
        \includegraphics[width=\textwidth]{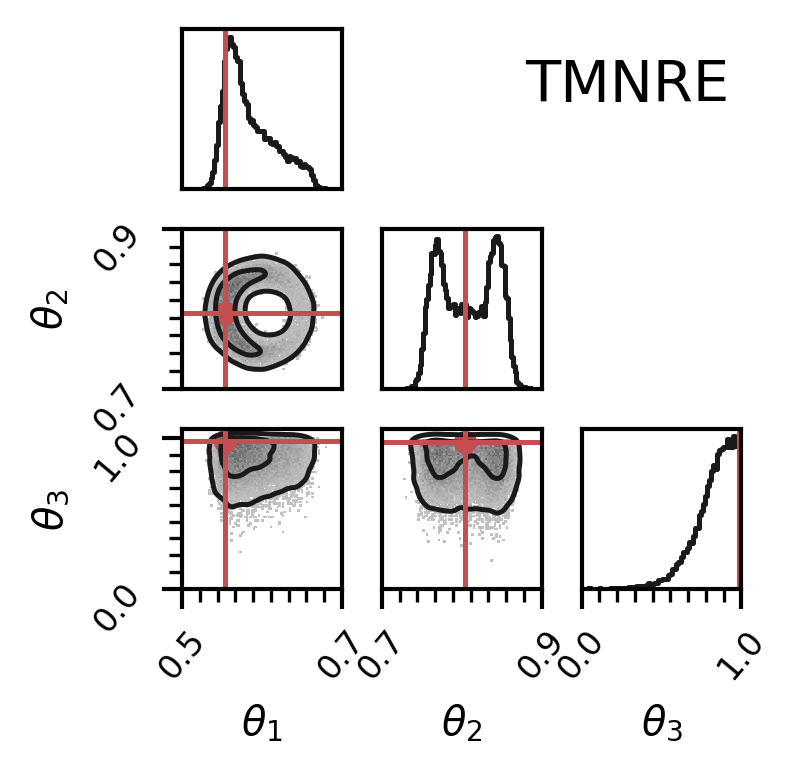}
    \end{minipage}\hfill
    \begin{minipage}{0.25\textwidth}
        \centering
        \includegraphics[width=\textwidth]{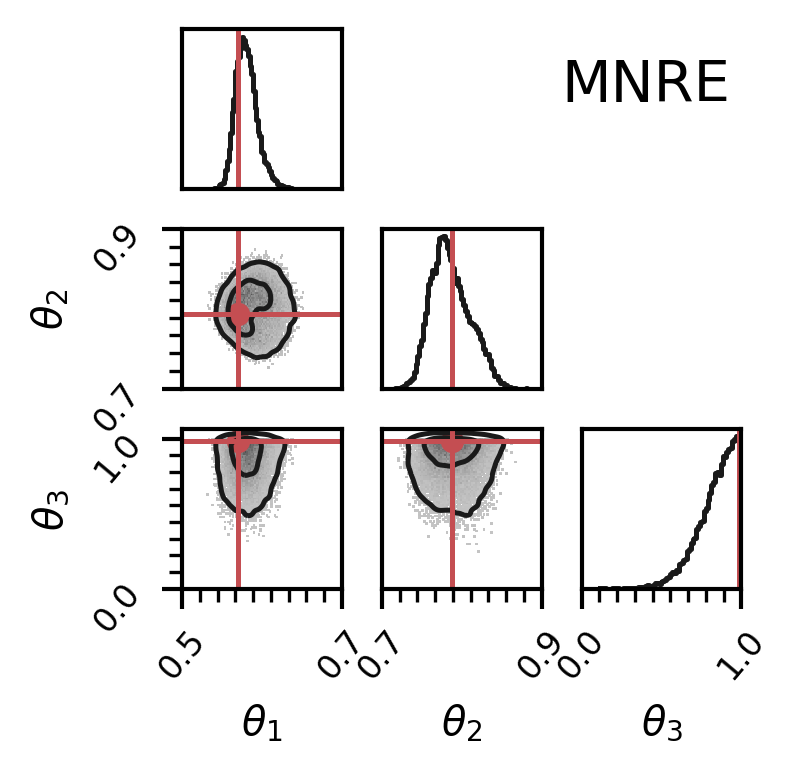}
    \end{minipage}\hfill
    \begin{minipage}{0.25\textwidth}
        \centering
        \includegraphics[width=\textwidth]{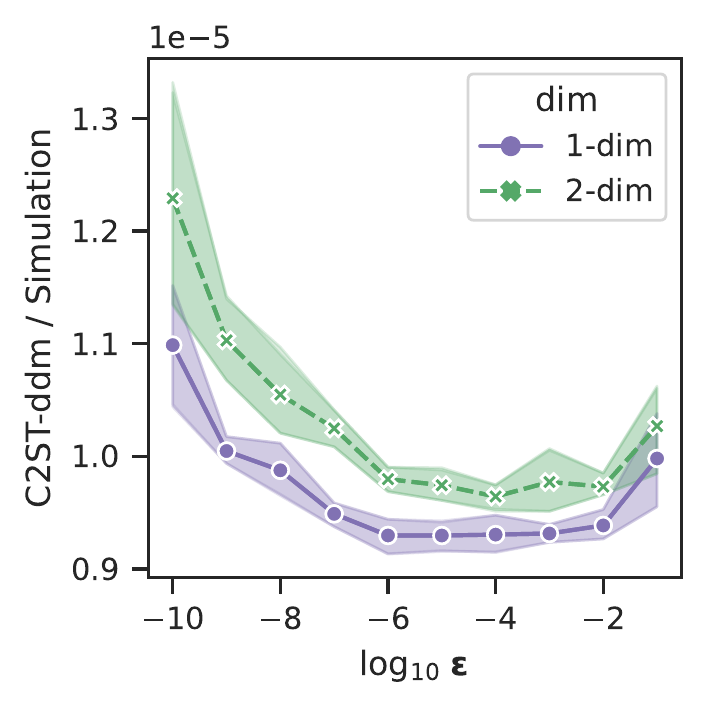}
    \end{minipage}\hfill
    \caption{%
        \textit{First through Third Panel:} View of torus marginal posteriors as estimated by rejection sampling, \TMNRE, and \MNRE. In some dimensions, the posterior extends the full unit cube prior width, while in others it is very narrow.  \TMNRE\ easily finds the asymmetric details after constraining to the relevant region while \MNRE\ does not. 
        \textit{Fourth Panel:} The results of a hyperparameter scan of $\epsilon$ on the torus task. The C2ST-ddm per simulation is reported versus $\epsilon$. The mean and 95\% confidence intervals are shown over five repetitions of the experiment. Lower values indicate better performance.
    }
    \label{fig:torus-posts-epsilon}
\end{figure}

We define a task which highlights the effectiveness of truncating the prior, namely a simulator with a very small torus shaped posterior. We present an ablation study of the truncation method along with a hyperparameter scan of $\epsilon$. Task details and additional experiments are presented in Appendix~\ref{apndx:experiments}.

\begin{wrapfigure}[27]{R}{0.35\textwidth}
	\begin{center}
    	\includegraphics[width=0.35\textwidth]{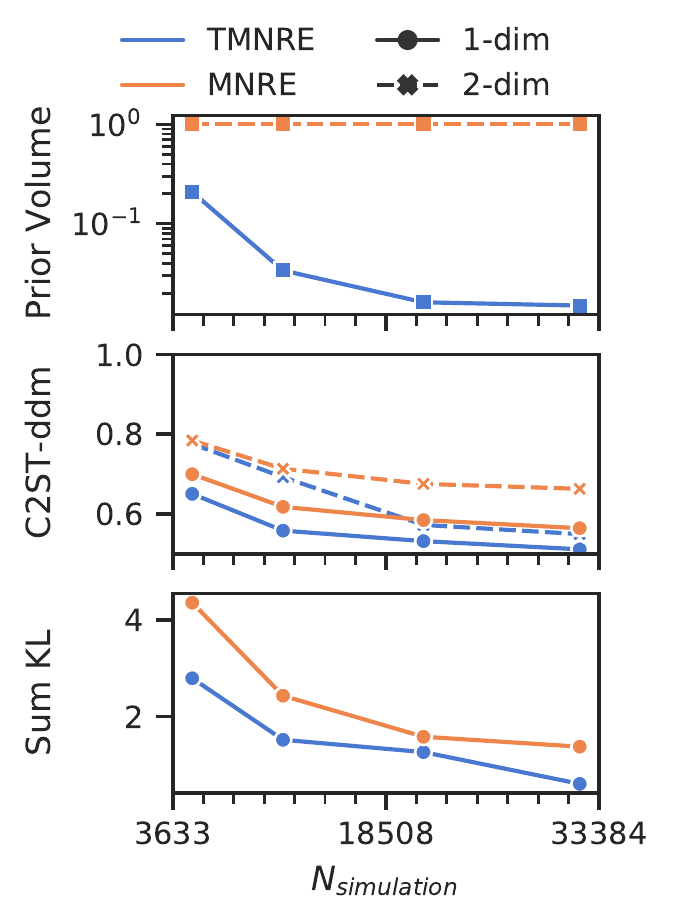}
	\end{center}
    \caption{
        Performance metrics on \MNRE\ and \TMNRE\ versus simulation budget. Budgets determined by truncation algorithm. \textit{T:} Prior volume. \textit{M:} According to C2ST-ddm, \TMNRE\ produces more accurate posteriors. \textit{L:} KL divergence summed over 1-dim marginals; same result.
	}
	\label{fig:torus-metrics}
\end{wrapfigure}

We ran Algorithm~\ref{alg:tmnre} which satisfied the stopping criterion after four rounds. We performed marginal likelihood-to-evidence ratio estimation on all one- and two-dimensional marginals for each step in the sequence of constrained regions, using the number of samples available that round. We also trained an estimator which used the same simulation budget but the samples were drawn from the unconstrained prior. We analyzed the prior volume, C2ST-ddm, and the sum of one dimensional KL divergences at each round for both methods. The posteriors are shown in Figure~\ref{fig:torus-posts-epsilon} and the performance metrics for the ablation study, are shown in Figure~\ref{fig:torus-metrics}.

\begin{figure}[hbt]
    \centering
    \begin{minipage}{0.26\textwidth}
        \centering
        \includegraphics[width=\textwidth]{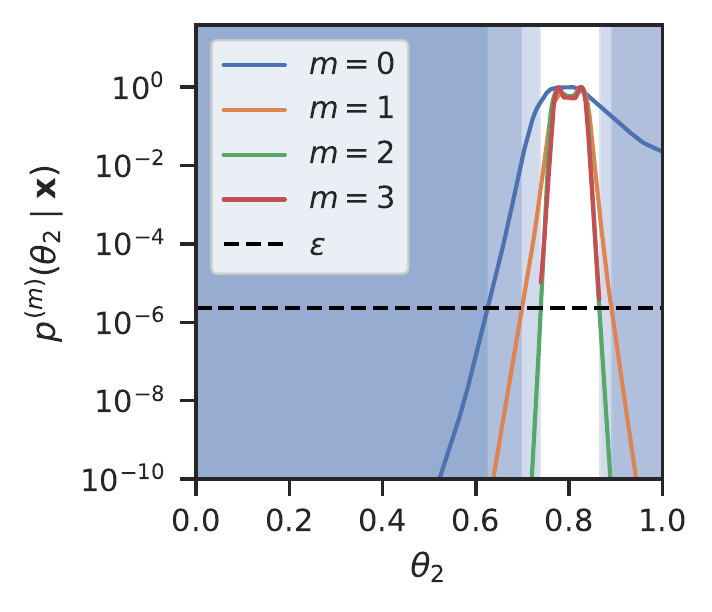}
    \end{minipage}
    \begin{minipage}{0.239\textwidth}
        \centering
        \includegraphics[width=\textwidth]{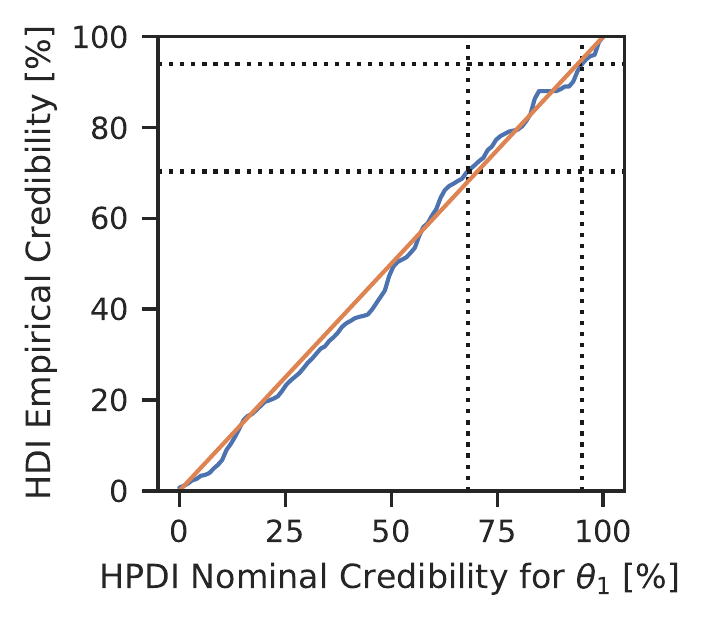}
    \end{minipage}
    \begin{minipage}{0.239\textwidth}
        \centering
        \includegraphics[width=\textwidth]{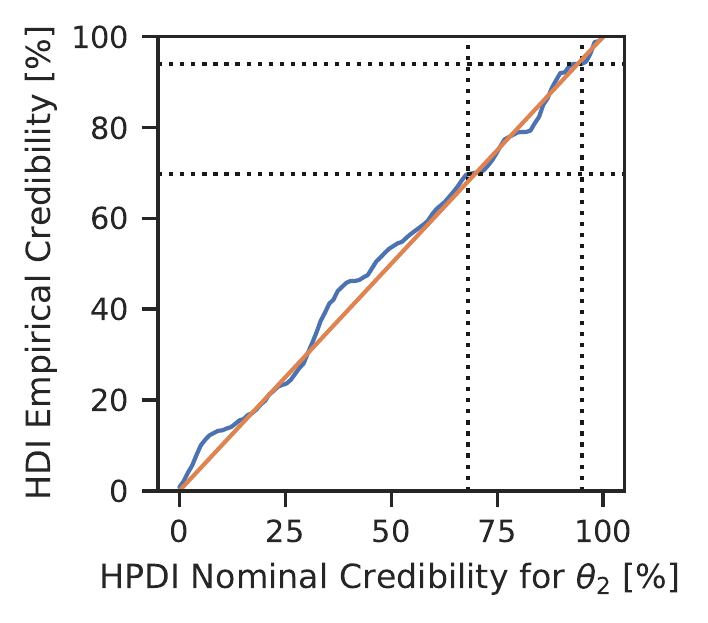}
    \end{minipage}
    \begin{minipage}{0.239\textwidth}
        \centering
        \includegraphics[width=\textwidth]{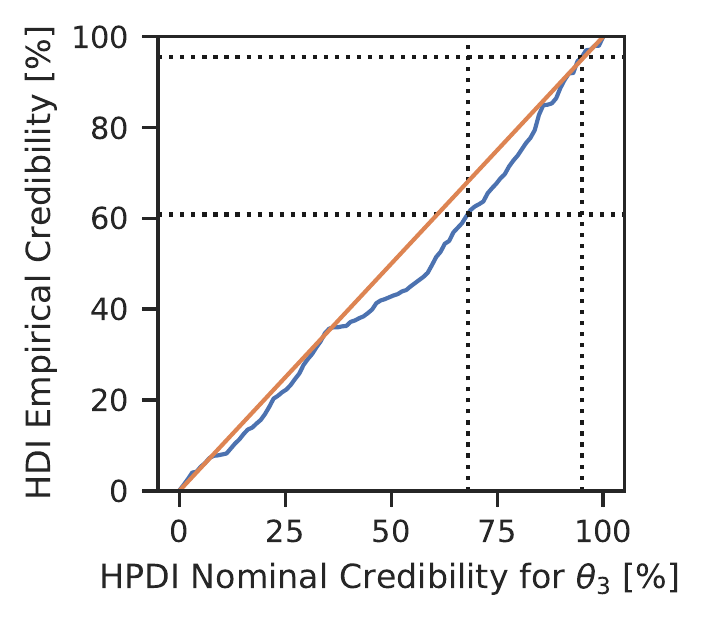}
    \end{minipage}
    \caption{%
        \textit{First panel:} The ratio estimator for $\theta_2$ in the 3-dim torus example, for several consecutive rounds. The estimator is conservative with limited data because each round's estimate converges from above to the final results (red line); i.e. the early rounds produce too-wide posteriors thus truncation decisions are made safely.  \textit{Second to Fourth Panels:}  Empirical versus nominal credibility for the highest posterior density intervals (HPDI) for each $\theta_{d}$ in the 3-dim torus. The line at (above) the diagonal implies accurate (conservative or wide) nominal credible intervals. See section \ref{sec:empirical_tests}.
    }
    \label{fig:checks}
\end{figure}

We found \TMNRE\ very accurately approximated all marginals at the maximum simulation budget. \MNRE\ placed mass in the correct region but missed the shape of the posterior entirely. 
\TMNRE\ improved simulation efficiency compared with \MNRE\ as indicated by the slope of the C2ST-ddm. The max-normalized posterior estimates at every round are plotted in Figure~\ref{fig:checks}. We note that given the limited training data in early rounds, our method predicts wider posteriors than the ground truth. These are called conservative posterior estimates and they are the preferred failure mode for practitioners \cite{hermans2021averting}. Other \SBI\ methods are tested on the torus in Appendix \ref{apndx:experiments}.

To determine the effects the hyperparameter $\epsilon$, we performed a grid search between $10^{-10}$ to $10^{-1}$ using \TMNRE\ on the same simulator. We reported the performance in terms of the C2ST-ddm per simulation in Figure~\ref{fig:torus-posts-epsilon} and repeated the experiment five times. We observe that the optimal value of $\epsilon$ was $10^{-6}$ since it was the most conservative value of $\epsilon$ that optimized the metric.

\subsection{Empirical tests of inference results through local amortization}
\label{sec:empirical_tests}

Our algorithm locally amortizes the posterior for parameters drawn from the constrained prior $p_\Gamma(\btheta)$. This opens the door for various experimental diagnostics to test the reliability of our trained inference networks with simulated data (which is also possible for \NRE~\cite{hermans2020towards}, but not for sequential methods that are exclusively targeting on one specific observation rather than a range of observations).  We demonstrate this by comparing the empirical credibility to the nominal credibility for the highest posterior density intervals, similarly to \cite{hermans2021averting} but we estimate using the truncated prior.

For the 3-dim torus example, we draw 10000 samples $(\bx, \theta_d) \sim p(\bx \mid \btheta) p_\Gamma(\btheta)$ from the truncated generative model.
For all samples, we generate marginal posteriors, $\hat p(\theta_d \mid \bx)$. For those marginal posteriors, we then derive the frequency with which $t\%$ highest density intervals contain the true value $\theta_d$.
The result is shown in Fig.~\ref{fig:checks}. It provides an immediate check of the reliability of our trained inference networks \emph{without knowing the ground truth}, and provides a safeguard against overconfident statements, which is critical for using the results of inference networks in a scientific context.

This empirical test is designed to show the consistency of the estimated nominal credible intervals across realizations of fabricated data but it does not address a generative model mismatch or access whether the estimated posterior corresponds to the ground truth.
When $\epsilon$ is small enough, the effects of truncation will not significantly impact this empirical test because of the accurate posterior estimate. Large $\epsilon$, that trim the tails of the estimated posterior aggressively, render this test unreliable because the estimated likelihood-to-evidence ratios will have inaccurate highest density intervals. It is possible to check whether $\epsilon$ is too large by observing a high-density posterior equicontour intersecting with a truncation bound, see Appendix~\ref{apndx:limitations}.

\subsection{Efficient marginal posteriors: the 10-dim eggbox}
\label{sec:eggbox}

We define a posterior that, when plotted in two dimensions, looks like a top-down view of an $2\times 2$ eggbox. Let $\btheta, \bg(\btheta) \in \mathbb{R}^{D}$ and $\theta_{k}$ denote the $k$th element of $\btheta$, then the simulator for this problem is defined $g_{k}(\btheta) = \sin(\theta_{k} \cdot \pi)$. To fix the posterior shape, we set $\theta_{k, o} = \frac{1}{4}, \, k = 1, 2, \ldots, D$ and $\bx_{0} = \bg(\btheta_o)$. 
The likelihood is determined by an additive noise model $p(\bx \mid \btheta) = \mathcal{N}(\bg(\btheta), \sigma^{2} \identity)$ with $\sigma = 0.1$. The total number of modes in our 10-dimensional model is $2^{10} = 1024$.  Realistic models do not typically feature such a regular mode structure, but this pattern enables comparison of the various algorithm's ability to handle multimodal data.

\begin{figure}[ht]
    \centering
    \begin{minipage}{0.19\textwidth}
        \centering
        \includegraphics[width=\textwidth]{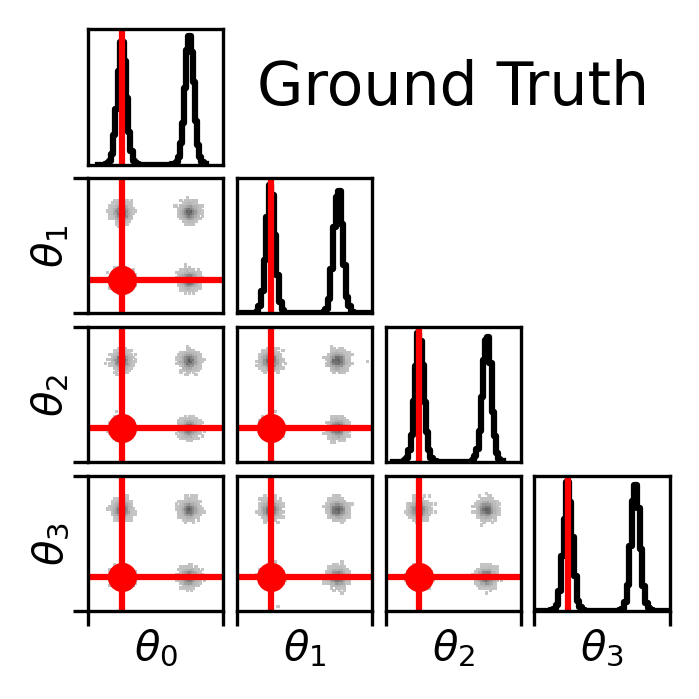}
    \end{minipage}
    \begin{minipage}{0.19\textwidth}
        \centering
        \includegraphics[width=\textwidth]{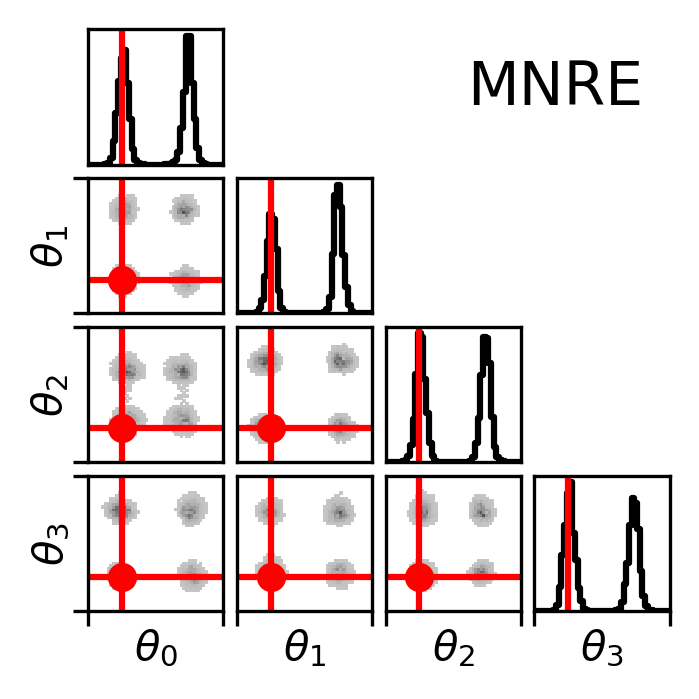}
    \end{minipage}
    \begin{minipage}{0.19\textwidth}
        \centering
        \includegraphics[width=\textwidth]{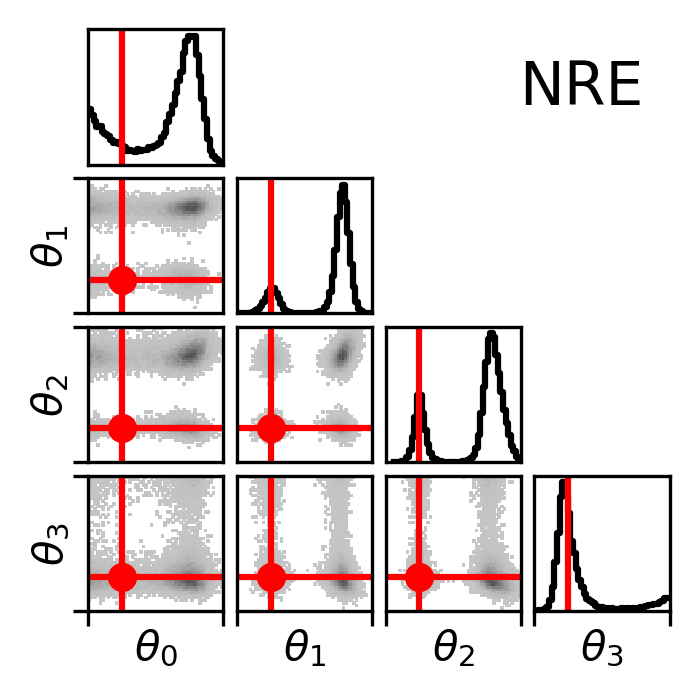}
    \end{minipage}
    \begin{minipage}{0.19\textwidth}
        \centering
        \includegraphics[width=\textwidth]{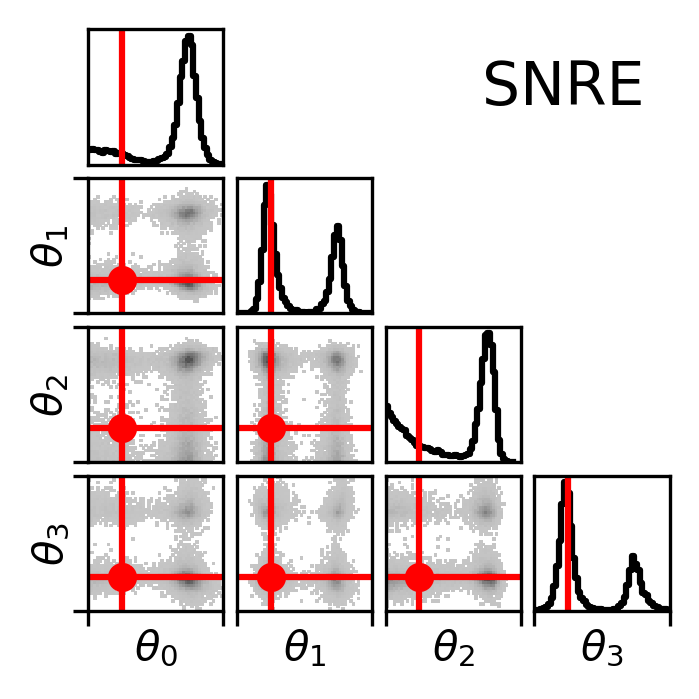}
    \end{minipage}
    \begin{minipage}{0.19\textwidth}
        \centering
        \includegraphics[width=\textwidth]{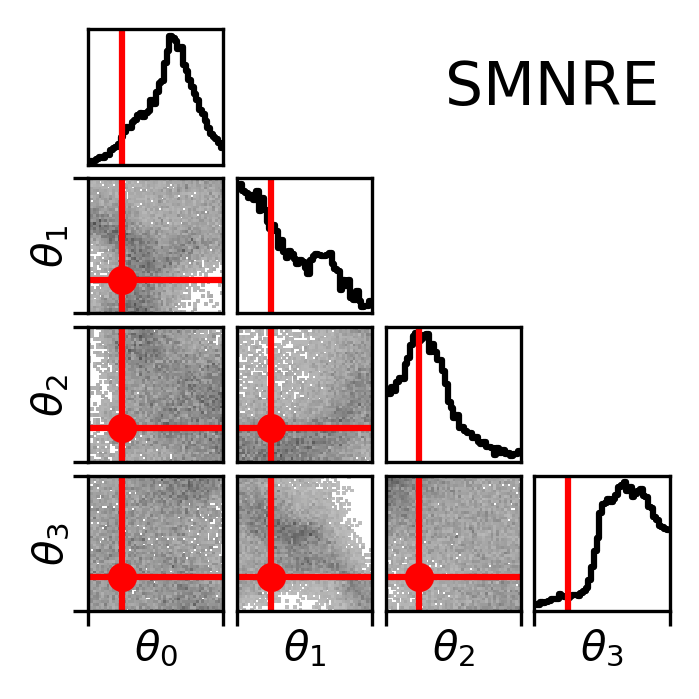}
    \end{minipage}
    \caption{%
        Posteriors from the 10-dim eggbox benchmark (only 4 parameters are shown for clarity). All methods received 10k training samples and produced 25k posterior samples. \NRE\ and \SNRE\ were trained jointly, while \MNRE\ and \SMNRE\ were trained marginally. \SMNRE\ cannot share training samples between marginals so each estimator received an equal share of the total simulation budget, 181 samples. Our method recovered the structure of the ground truth, unlike \NRE, \SNRE, or \SMNRE.
    }
    \label{fig:eggbox-corner}
\end{figure}

Given 10,000 training samples drawn from the prior and a $D=10$ dimensional parameter space, we trained \MNRE\ to estimate all one- and two-dimensional marginals, the \SBI~\cite{sbi} implementation of \NRE\ and \SNRE\ on the joint, and finally a marginalized version of \SNRE\ (\SMNRE). In \SMNRE, we divided the samples across 10 rounds and each round proposed samples according to the previous round's posterior distribution for the predicted marginals, but the initial prior for the nuisance parameters. Since, in a general setting, \SMNRE\ cannot use samples from another marginal estimator, we divided the 10,000 training samples evenly among the 55, one- and two-dimensional marginal estimators, each estimator receiving 181 training samples. 25,000 samples from each reported posterior are visible in Figure~\ref{fig:eggbox-corner}. \TMNRE\ recovered the structure of the ground truth marginal posteriors, providing empirical evidence that estimating marginals directly can provide high accuracy at low simulation budgets for complex high-dimensional posteriors. Experiments using other \SBI\ methods are in Appendix \ref{apndx:experiments} along with discussion of a rotated version of the problem.

\subsection{Practitioner's use case: cosmological inference with a simulator}
\label{sec:main_text_physics_example}

Parameter inference is commonplace in astrophysics and cosmology. We are interested in a simulator with six parameters that returns three angular power spectra, as they would be measured by an idealized Cosmic Microwave Background (CMB) experiment. A budget of 5,000 simulations is sufficient for \MNRE\ to accurately estimate the corner plot, while \NRE\ produces unconstrained and inaccurate marginal posterior estimates. Due to space constraints, we could not include this practical application of \MNRE\ in the body of this paper. Please see Appendix \ref{apndx:cosmology} for details of the simulator, the inference technique, and the results. A complete study of the topic is currently in preparation \cite{our_cosmology_paper}.

\section{Discussion and conclusions}
\label{sec:conclusions}

We presented Truncated Marginal Neural Ratio Estimation (\TMNRE), a simulation-based inference algorithm based on \NRE.  
The core idea of our algorithm is to focus on most probable parameter regions by truncating marginal posteriors in their very low-probability tails. For Gaussian posteriors this is typically beyond $5\sigma$ and does not significantly affect the higher density contours.
In addition to performing on par or better than existing algorithms on standard benchmarks, \TMNRE\ is better suited to the practitioner's needs than other algorithms because it offers simulation efficient marginal posterior estimation and the capacity to perform efficient consistency checks through local amortization. These features are particularly desirable to scientists whose simulators are expensive and rife with nuisance parameters.

\TMNRE\ uses a sequence of training and sampling rounds to automatically produce parameters with high posterior density, i.e. relevant to a particular observation $\bx_{o}$. The output of this sequence is a hyperrectangular approximation to a highest posterior density region, implicitly defined by hyperparameter $\epsilon$. 
That implies parameters from within this constrained region are more likely to produce data similar to $\bx_{o}$ than simulations from outside the region.
Using data drawn from this constrained region, \TMNRE\ estimates any marginal posterior of interest directly using a marginal likelihood-to-evidence ratio; a simpler and more practical technique than estimating the entire joint posterior. 
Finally, by construction, our targeted inference method can accurately estimate posteriors of simulations from within the constrained parameter region. This freedom facilitates fast empirical studies of the nominal credible regions, which are of critical importance in real-life applications when there is no ground truth posterior to compare to.

On the \SBI\ benchmark \cite{sbibm}, we found that \TMNRE\ is on par with the most effective \SBI\ algorithms, such as \SNRE\ \cite{Hermans2019}, as measured by the C2ST performance metric, Fig.~\ref{fig:sbibm}.  We highlighted the benefits of \TMNRE\ using two showcase tasks: a torus-shaped posterior and a multimodal eggbox-shaped posterior. The torus featured a very narrow posterior that \TMNRE\ found and accurately learned while simple \MNRE\ failed to do so, Fig.~\ref{fig:torus-metrics}. We demonstrated validity of our iterative procedure, and the ability to perform important validation tests by testing the nominal credible intervals empirically, Fig.~\ref{fig:checks}. The eggbox's joint posterior was challenging for \NRE-based methods but \MNRE\ efficiently estimated the marginal posteriors, Fig.~\ref{fig:eggbox-corner}. Further torus and eggbox experiments are in Appendix \ref{apndx:experiments} and a cosmology posterior is estimated in Appendix \ref{apndx:cosmology}.

The presented algorithm is aimed at marginal posterior inference, which is a typical goal for scientific applications, but does not allow, e.g., to evaluate the posterior predictive distribution which requires the joint posterior.  Furthermore, our algorithm particularly shines for
high-dimensional problems with complex and/or narrow posteriors, whereas we expect that simpler problems may be better suited to other \SBI\ methods. We address further limitations of our method in Appendix~\ref{apndx:limitations}. 

We note that the hyperrectangular indicator function, defined in Eq.~\eqref{eqn:GammaDef}, is not optimal if some of the parameters are strongly correlated.  However, it can be straightforwardly extended to more complex shapes. The challenge is to efficiently define the boundaries of the indicator function and sample from within it, a problem tackled by effective nested sampling algorithms~\cite{dynesty}.

This work is primarily foundational and the societal impacts, other than the cost of training machine learning models, would therefore be drawn from a hypothetical application. As this is an inference method, it would be possible to apply it to biased simulators which could reinforce unethical patterns.
In general, the societal impacts are closely tied to the implications of the simulators themselves.

\cleardoublepage

\begin{ack}
	This work uses \texttt{numpy} \cite{harris2020array}, \texttt{scipy} \cite{2020SciPy-NMeth}, \texttt{seaborn} \cite{Waskom2021}, \texttt{matplotlib} \cite{Hunter:2007}, \texttt{altair} \cite{VanderPlas2018, Satyanarayan2017}, \texttt{pandas} \cite{reback2020pandas, mckinney-proc-scipy-2010} \texttt{pytorch} \cite{pytorch}, and \texttt{jupyter} \cite{jupyter}. 
	Benjamin Kurt Miller is funded by the University of Amsterdam Faculty of Science (FNWI), Informatics Institute (IvI), and the Institute of Physics (IoP).
	
	We want to thank the DAS-5 computing cluster for access to their TitanX GPUs. DAS-5 is funded by the NWO/NCF (the Netherlands Organization for Scientific Research).
	We received funding from the European Research Council (ERC) under the European Union’s Horizon 2020 research and innovation programme (Grant agreement No. 864035 -- UnDark).
\end{ack}

\bibliographystyle{unsrt}
\bibliography{bibliography}

\cleardoublepage

\appendix

\section{Experiments}
\label{apndx:experiments}

In this section we present the relevant experimental details including a comprehensive list of experiments in Table~\ref{table:experiments}. We first discuss the computational setting and approximate computational cost. Afterwards the details for the \SBI\ benchmark, the torus, and the eggbox are presented along with additional experiments for those problems. Finally we discuss how we generate our datasets and how we can use the estimated likelihood-to-evidence ratio in practice. In all tests, we applied \TMNRE\ with the hyperparameters laid out in Table~\ref{table:hyperparameters}.

\begin{table}[htb]
	\caption{Experiments}
	\label{table:experiments}
	\centering
	\resizebox{\textwidth}{!}{
        \begin{tabular}{lcccccccc}
            \toprule
            Task & \shortstack{Parameter \\ Dimension} & Algorithm & Simulations & \shortstack{Trained \\ Marginalization} & \shortstack{Evaluated \\ Marginalization} & Constrained & Metric \\
            \midrule
            Two Moons & 2 & \TMNRE & $\{1, 10, 100\}\textrm{E}3$ & 1d, 2d & 1d, 2d & 1d & C2ST-ddm \\
            Two Moons & 2 & \SBI & $\{1, 10, 100\}\textrm{E}3$ & Joint & 1d, 2d & & C2ST-ddm \\
            Gaussian Linear & 10 & \TMNRE & $\{1, 10, 100\}\textrm{E}3$ & 1d, 2d & 1d, 2d & 1d & C2ST-ddm \\
            Gaussian Linear & 10 & \SBI & $\{1, 10, 100\}\textrm{E}3$ & Joint & 1d, 2d & & C2ST-ddm \\
            Gaussian Linear Uniform & 10 & \TMNRE & $\{1, 10, 100\}\textrm{E}3$ & 1d, 2d & 1d, 2d & 1d & C2ST-ddm \\
            Gaussian Linear Uniform & 10 & \SBI & $\{1, 10, 100\}\textrm{E}3$ & Joint & 1d, 2d & & C2ST-ddm \\
            SLCP & 5 & \TMNRE & $\{1, 10, 100\}\textrm{E}3$ & 1d, 2d & 1d, 2d & 1d & C2ST-ddm \\
            SLCP & 5 & \SBI & $\{1, 10, 100\}\textrm{E}3$ & Joint & 1d, 2d & & C2ST-ddm \\
            SLCP (distractors) & 5 & \TMNRE & $\{1, 10, 100\}\textrm{E}3$ & 1d, 2d & 1d, 2d & 1d & C2ST-ddm \\
            SLCP (distractors) & 5 & \SBI & $\{1, 10, 100\}\textrm{E}3$ & Joint & 1d, 2d & & C2ST-ddm \\
            Gaussian Mixture & 2 & \TMNRE & $\{1, 10, 100\}\textrm{E}3$ & 1d, 2d & 1d, 2d & 1d & C2ST-ddm \\
            Gaussian Mixture & 2 & \SBI & $\{1, 10, 100\}\textrm{E}3$ & Joint & 1d, 2d & & C2ST-ddm \\
            SIR & 2 & \TMNRE & $\{1, 10, 100\}\textrm{E}3$ & 1d, 2d & 1d, 2d & 1d & C2ST-ddm \\
            SIR & 2 & \SBI & $\{1, 10, 100\}\textrm{E}3$ & Joint & 1d, 2d & & C2ST-ddm \\
            Lotka-Volterra & 4 & \TMNRE & $\{1, 10, 100\}\textrm{E}3$ & 1d, 2d & 1d, 2d & 1d & C2ST-ddm \\
            Lotka-Volterra & 2 & \SBI & $\{1, 10, 100\}\textrm{E}3$ & Joint & 1d, 2d & & C2ST-ddm \\
            \midrule
            Torus & 3 & \TMNRE & 4985, 11322, 21127, 32032 & 1d, 2d & 1d, 2d & 1d & C2ST-ddm, KLD, Visual \\
            Torus & 3 & \MNRE & 4985, 11322, 21127, 32032 & 1d, 2d & 1d, 2d &  & C2ST-ddm, KLD, Visual \\
            Torus & 3 & \SBI & 4985, 11322, 21127, 32032 & Joint &  &  & Visual \\
            \midrule
            Torus (epsilon scan) & 3 & \TMNRE & $\sim 30~\textrm{E}3$ & 1d, 2d & 1d, 2d & 1d & C2ST-ddm / simulation \\
            \midrule
            Egg Box 2 modes / dim & 10 & \MNRE & $10~\textrm{E}3$ & 1d, 2d &  & & Visual \\
            Egg Box 2 modes / dim & 10 & \NRE & $10~\textrm{E}3$ & Joint &  & & Visual \\
            Egg Box 2 modes / dim & 10 & \SNRE & $10~\textrm{E}3$ & Joint &  & & Visual \\
            Egg Box 2 modes / dim & 10 & \SMNRE & $10~\textrm{E}3$ & 1d, 2d &  & & Visual \\
            Egg Box 2 modes / dim & 10 & Remaining \SBI & $10~\textrm{E}3$ & Joint & & & Visual \\
            \midrule
            Rotated Egg Box & 10 & \MNRE & $10~\textrm{E}3$ & 1d, 2d &  & & Visual \\
            Rotated Egg Box & 10 & \SBI & $10~\textrm{E}3$ & Joint & & & Visual \\
            \bottomrule
        \end{tabular}
    }
\end{table}

\begin{table}[htb]
	\caption{\TMNRE\ Hyperparameters}
	\label{table:hyperparameters}
	\centering
    \begin{tabular}{ll}
        \toprule
        Hyperparameter & Value \\
        \midrule
        Activation Function & \textsc{relu} \\
        \textsc{amsgrad} & No \\
        Architecture & \textsc{resnet} (2 blocks) \\
        Batch normalization & Yes \\
        Batch size & 128 \\
        Criterion & \textsc{bce} \\
        Dropout & No \\
        Early stopping patience & 20 \\
        $\epsilon$ & $e^{-13} \approx 10^{-6}$ \\
        Hidden features & 64 \\
        Percent validation & 10\% \\
        Reduce lr factor & 0.1 \\
        Reduce lr patience & 5 \\
        Max epochs & 300 \\
        Max rounds & 10 \\
        Learning rate & 0.01 \\
        Learning rate scheduling & Decay on plateau \\
        Optimizer & \textsc{adam} \\
        Weight Decay & 0.0 \\
        Standard-score Observations & online \\
        Standard-score Parameters & online \\
        \bottomrule
    \end{tabular}
\end{table}

\subsection{Total Compute}
\label{apndx:compute}
Most calculations were performed on a local computing cluster which offered ten TitanX GPU nodes. We estimate the total computation time, including prototype runs, was approximately 968 GPU hours. We calculated the cost of one run of the benchmark then multiplied it by 10 for this estimation. The computation of the C2ST-ddm on the marginals from existing data was performed on the same cluster but using cpu nodes. According to \texttt{mlco2.github.io} this would imply 104.54 kg CO$_2$ at a normal institution; however, our cluster is run exclusively on wind power.

\subsection{SBIBM details}

\begin{table}[htb]
	\caption{Actual bounds of stochastic simulation budget for \TMNRE\ along with number of rounds before the stoppping criterion was reached. Maximum of one round implies that there was no truncation and the method is effectively doing \MNRE.}
	\label{table:tmnre_num_simulations}
	\centering
	\resizebox{\textwidth}{!}{
    \begin{tabular}{llrrrr}
    \toprule
              &        &  Min simulation count &  Max simulation count &  Min Rounds &  Max Rounds \\
    task & num\_simulations &                       &                       &             &             \\
    \midrule
    gaussian\_linear & 1000   &                   960 &                  1065 &           1 &           3 \\
              & 10000  &                  9966 &                 12004 &           3 &           7 \\
              & 100000 &                 99702 &                115784 &           3 &           6 \\
    gaussian\_linear\_uniform & 1000   &                   952 &                  1056 &           1 &           1 \\
              & 10000  &                  9760 &                 10469 &           1 &           4 \\
              & 100000 &                100223 &                105468 &           1 &           4 \\
    gaussian\_mixture & 1000   &                   954 &                  1072 &           1 &           4 \\
              & 10000  &                  9902 &                 10582 &           2 &           4 \\
              & 100000 &                 99567 &                105704 &           2 &           3 \\
    lotka\_volterra & 1000   &                   966 &                  1051 &           1 &           2 \\
              & 10000  &                  9824 &                 11916 &           1 &           6 \\
              & 100000 &                 99791 &                128290 &           2 &           5 \\
    sir & 1000   &                   951 &                  1024 &           1 &           5 \\
              & 10000  &                  9973 &                 10128 &           2 &           5 \\
              & 100000 &                 99611 &                100547 &           2 &           3 \\
    slcp & 1000   &                   949 &                  1050 &           1 &           1 \\
              & 10000  &                  9901 &                 10546 &           1 &           1 \\
              & 100000 &                 99616 &                104968 &           1 &           2 \\
    slcp\_distractors & 1000   &                   951 &                  1035 &           1 &           1 \\
              & 10000  &                  9931 &                 10141 &           1 &           1 \\
              & 100000 &                 99431 &                100882 &           1 &           1 \\
    two\_moons & 1000   &                   934 &                  1056 &           1 &           4 \\
              & 10000  &                  9941 &                 10558 &           1 &           4 \\
              & 100000 &                 99863 &                104919 &           2 &           3 \\
    \bottomrule
    \end{tabular}
    }
\end{table}

We performed a marginalized version of the \SBI\ benchmark on all tasks from \cite{sbibm}, except the Bernoulli Generalized Linear Model. Each task has ten parameters drawn from the corresponding prior. Each of those parameters are pushed through the simulator and those become ten observations with a known ground truth posterior and true generating parameter.  For the details of each task we refer the reader to Ref. \cite{sbibm} where they are defined at great length. A summary of some of the details for each of these tasks, and the algorithm applied to them, are contained in the Experiment Table~\ref{table:experiments}. 

Although it is not required for \TMNRE, we applied the stochastic process from \cite{swyft} to generate samples from the prior distribution. This led to an estimated number of samples for each task. In general, the number of training samples lied within $\sim 5\%$ of reported value in Figure \ref{fig:sbibm}. This was not true for the Lotka-Volterra and Gaussian Linear tasks because some runs had more truncation rounds than expected with every round introducing more samples.

Five runs did not converge with \TMNRE, those runs were on Gaussian Linear at 1,000 simulations with observation numbers 4 and 5. Lotka-Volterra had the same problem with 10,000 for observations 2 and 6 and again at 100,000 simulations with observation number 6.

The other methods estimated the joint posterior in some manner while \TMNRE\ targeted the marginals directly. The full list of alternative methods are called \REJABC, \NLE, \NPE, \NREB, \SMCABC, \SNLE, \SNPE, \SNREB. These methods represent a significant portion of the neural simulation-based inference literature and will not be described in detail here. Please consult Ref. \cite{sbibm}.

We defined a summary of the C2ST across the task's marginals by taking the average over same-dimensional marginals and averaged over the observations, see \eqref{eqn:c2st-ddm}. These values are reported for all methods in Figure~\ref{fig:sbibm} where the 95\% confidence intervals are computed for the C2ST-ddm over observations, i.e. the variance across marginals in the C2ST-ddm calculation is not carried forward into the reported uncertainty. We found it to be very small compared to the reported values and was unlikely to make a significant difference.

The authors note that data and code was used from \textsc{sbibm} which can be found on GitHub at \texttt{https://github.com/sbi-benchmark/sbibm}. It is distributed with the MIT license.

\paragraph{Our method}
\TMNRE\ was trained to learn all one- and two-dimensional likelihood-to-evidence ratios thereby predicting the posterior distribution. Since we applied \TMNRE, the algorithm truncated the prior distribution depending on the learned marginal likelihood-to-evidence ratio. We gave a generous maximum of ten rounds but no task used so many. The maximum was seven before the stopping criterion was satisfied. We used the ratio of the constrained prior mass from the current round to the previous round, namely $\beta$ in Algorithm~\ref{alg:tmnre}, as a stopping criterion and set it to 0.8. The stopping criterion was satisfied after a certain number of rounds details about the maximum and minimum round for every task, at every budget, can be found in Table~\ref{table:tmnre_num_simulations}. We applied the heuristic for the simulation budget found in Appendix~\ref{apndx:data_gen}.

The final estimated likelihood-to-evidence ratio approximates the posterior on the constrained region. Samples were drawn from this posterior using rejection sampling. The samples from these marginals, in the constrained region, are used for the reported C2ST-ddm in Figure~\ref{fig:sbibm}.

\subsection{Torus details}
We use a simulator and prior with a torus shaped posterior to showcase three aspects of \TMNRE. The ground truth can be seen on the left in Figure~\ref{fig:torus-posts-epsilon}. We proceed with the details of the simulator followed but subsections which give the details for every showcase experiment.

If we let $\btheta, \bg(\btheta) \in \mathbb{R}^{3}$ and $\theta_{k}$ denote the $k$th element of $\btheta$, then the simulator for this problem is defined $\bg(\btheta) = (\theta_{0}, \sqrt{(\theta_{0}-a)^2+(\theta_{1}-b)^2}, \theta_{2})^{T}$. The likelihood is defined by an additive noise model, namely $p(\bx \mid \btheta) = \mathcal{N}\left(g(\btheta), \bSigma \right)$ where $a, b$ are constant scalars and $\bSigma$ is a diagonal, positive definite matrix. 
In our experiments we let $a = 0.6$, $b = 0. 8$, and $\bSigma = \textrm{diag}(0.03^{2}, 0.005^{2}, 0.2^{2})$. To ensure an approximately torus-shaped posterior, we select a ``noiseless'' observation of interest $\bx_{0} = \bg(\btheta_{0})$ and parameters $\btheta_{0} = (0.57, 0.8, 1.0)^{T}$. 

\subsubsection{Torus TMNRE and MNRE Metrics}

Since we did not have a clear simulation budget during the initial run of \TMNRE, we determined the number of simulations in the following round by multiplying the retained simulations by 1.5 and sampling from a Poisson distribution. We started with 5,000 requested samples and up to 10 rounds. In the end that meant we ran Algorithm~\ref{alg:tmnre} with the following number of simulations in each round: 4985, 11322, 21127, 32032. The stopping criterion was met in four rounds, before the maximum number of rounds was reached. 

\begin{figure}[hbt]
	\centering
	\includegraphics[width=1.0\textwidth]{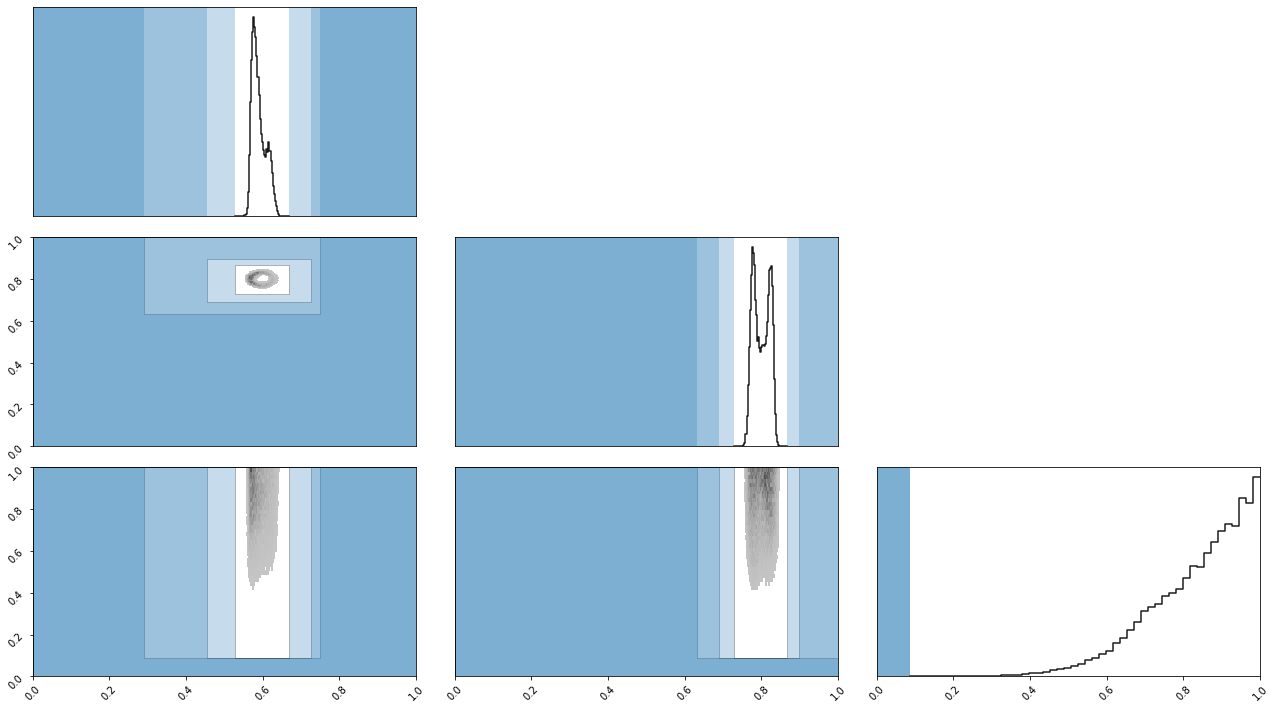}
	\caption{
	    An example of what the truncation process looks like in both one and two dimensions. At each round, the truncated region of the indicator function is denoted in blue. The region is slightly transparent so that the evolution over rounds can be discerned. This plot helps visualize that truncation only occurs at very, very low posterior density. The entire prior region is shown to emphasize that naive sampling results in few samples within the region of interest.
    }
	\label{fig:torus-corner-conceptual}
\end{figure}

A sample visualization of this truncation process is visible in Figure~\ref{fig:torus-corner-conceptual}. As described in the text, each of these truncated priors were utilized for an ablation study where we estimated the marginal likelihood-to-evidence ratio using either the truncated prior or the true prior. In effect, testing the value of \TMNRE\ versus \MNRE. Once the number of simulations were fixed by \TMNRE\ we used exactly the same number of simulations at that stage with \MNRE.

\subsubsection{Epsilon Hyperparameter Scan}

To determine a useful default value for the cutoff threshold $\epsilon$, we ran \TMNRE\ on the torus simulator, as described above, at 10 different values of epsilon. Namely, $\epsilon_{i} \in \{10^{i} : i = -1, \ldots, -10 \}$. At every round, the simulator requested approximately 10,000 more simulations than were retained from the previous round. The amount of simulations was determined stochastically, see Appendix~\ref{apndx:data_gen}. Once the method had hit the stopping criteria, the one- and two-dimensional C2ST-ddm was computed and normalized by the number of simulations required to generate it. The results were plotted by truncation cutoff $\epsilon$ on the right in Figure~\ref{fig:torus-posts-epsilon}. We determined that $10^{-6}$ minimized the C2ST-ddm approximately as well as the global minima $10^{-4}$ while truncating the prior more conservatively.

\subsubsection{Empirical Self-Consistency Test}

When we do not have a ground truth to compare to, it is important that we can determine whether the nominal credible intervals correspond to the true credible intervals. We propose to do so by comparing the nominal credibility to the empirical credibility. An explanation of how to calculate this performance metric is provided in Appendix~\ref{apndx:ctest}.

The authors note, as addressed briefly in the main text, that our consistency test checks the calibration \cite{hermans2021averting} of the posterior to the \emph{truncated} prior, rather than the true prior. This may introduce a bias into the estimation of the self-consistency compared with drawing samples from the truth prior, although we expect it to be small when $\epsilon$ is small. Investigation of the effects are left for future work.

\subsection{Alternative Simulation-based Inference Methods on the Torus}

\begin{figure}[bth]
    \centering
    \begin{minipage}{0.24\textwidth}
        \centering
        \includegraphics[width=\textwidth]{torus/torus-ref-corner}
    \end{minipage}\hfill
    \begin{minipage}{0.24\textwidth}
        \centering
        \includegraphics[width=\textwidth]{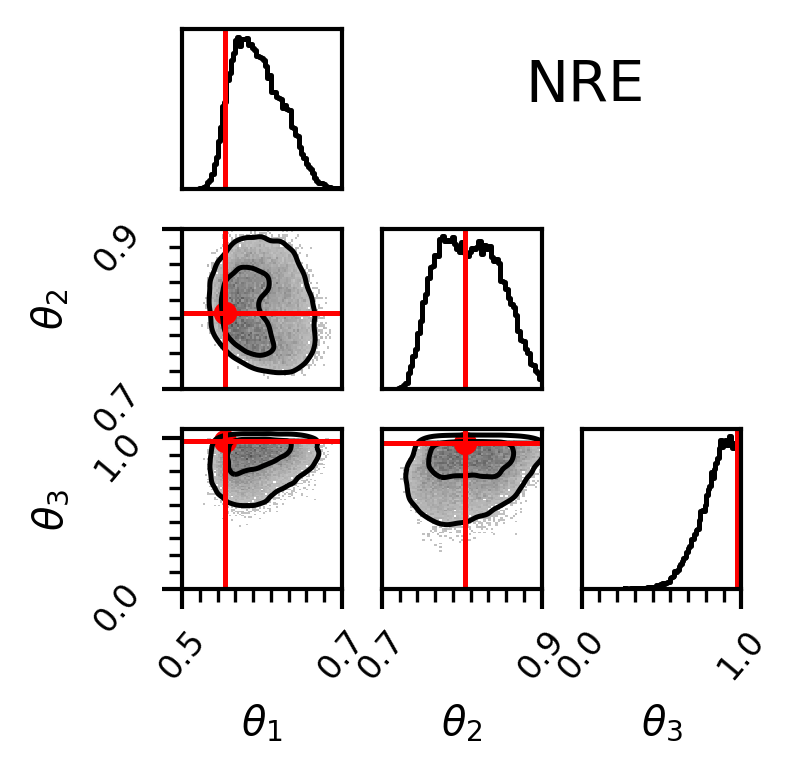}
    \end{minipage}\hfill
    \begin{minipage}{0.24\textwidth}
        \centering
        \includegraphics[width=\textwidth]{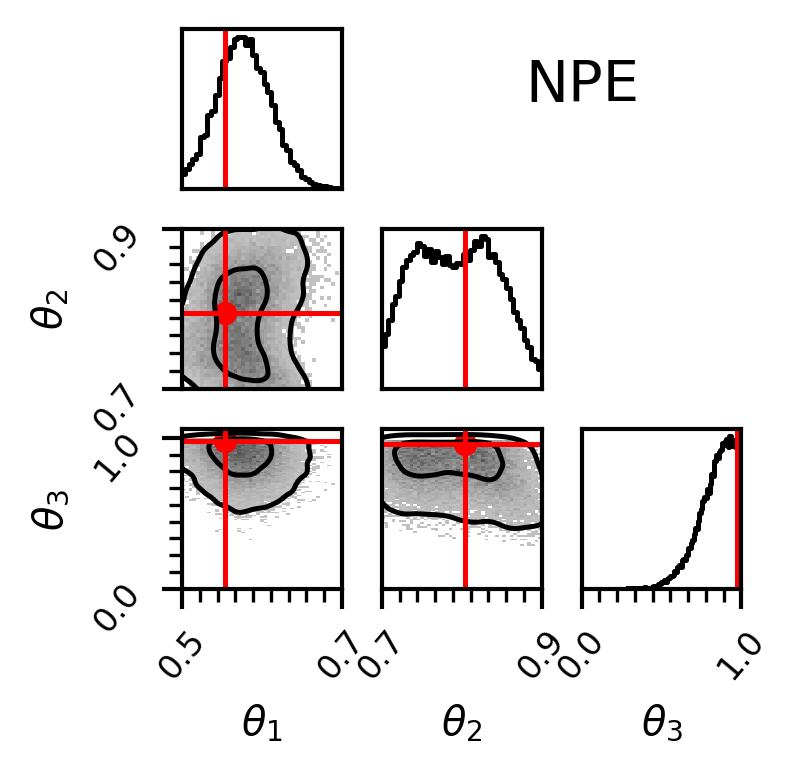}
    \end{minipage}\hfill
    \begin{minipage}{0.24\textwidth}
        \centering
        \includegraphics[width=\textwidth]{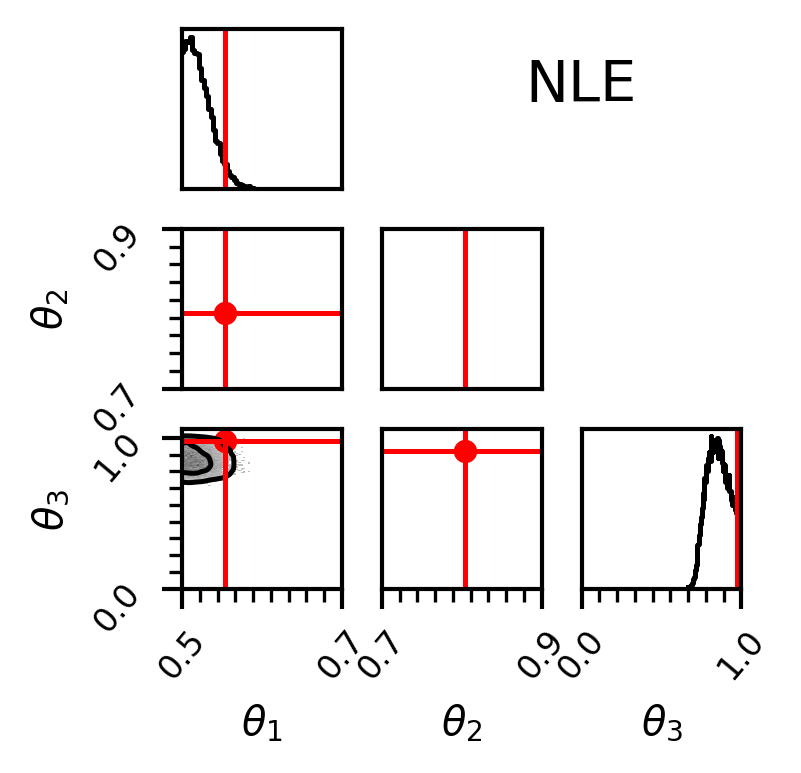}
    \end{minipage}\hfill
    \begin{minipage}{0.24\textwidth}
        \hfill
    \end{minipage}\hfill
    \begin{minipage}{0.24\textwidth}
        \centering
        \includegraphics[width=\textwidth]{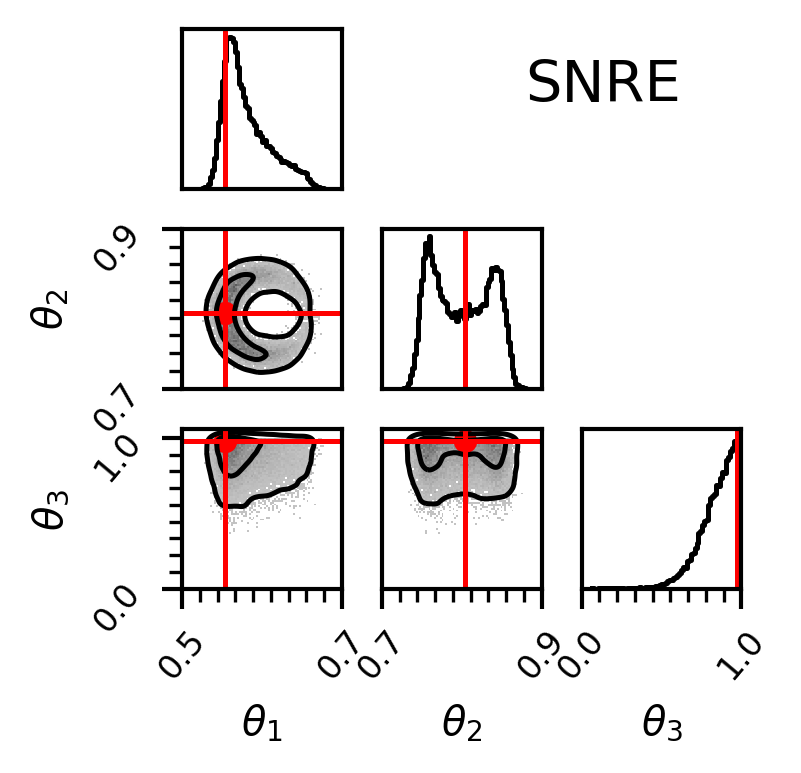}
    \end{minipage}\hfill
    \begin{minipage}{0.24\textwidth}
        \centering
        \includegraphics[width=\textwidth]{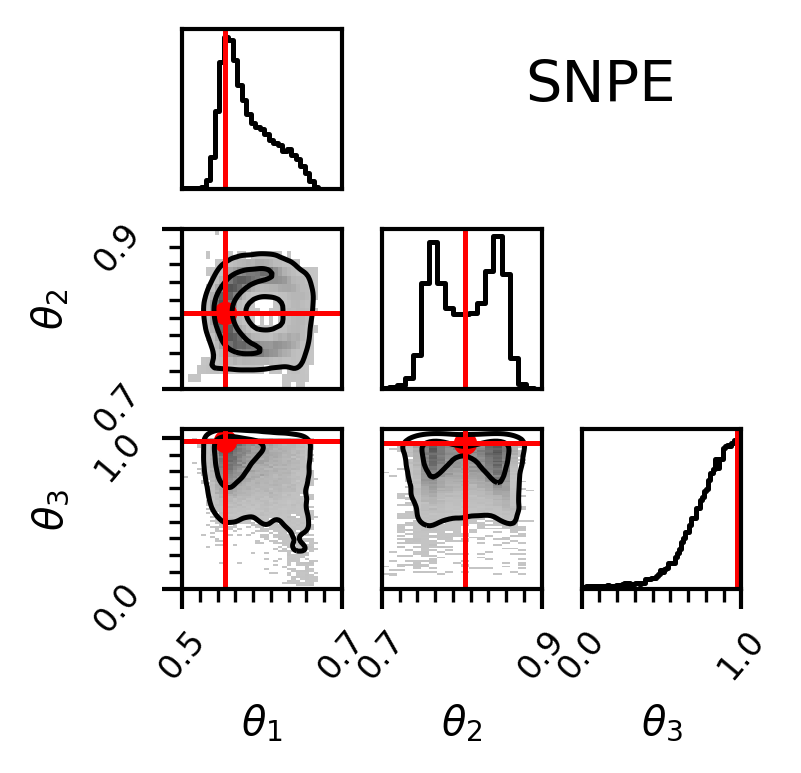}
    \end{minipage}\hfill
    \begin{minipage}{0.24\textwidth}
        \centering
        \includegraphics[width=\textwidth]{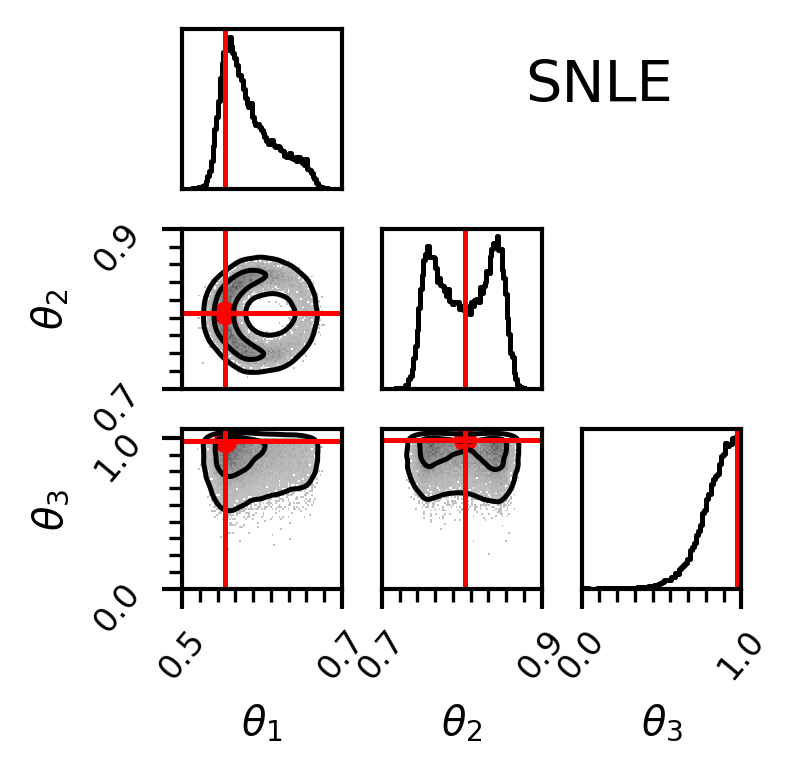}
    \end{minipage}\hfill
    \caption{%
        Joint posteriors estimated by various popular \SBI\ methods on the torus problem. When compared to the prior, the narrowness of this posterior renders most untargeted simulations uninformative. Only sequential or truncation-based methods are able to recover accurate estimates at the presented simulation budget of 32032.
        \textit{Left column}: Torus ground truth from rejection sampling.
        \textit{Mid-left column}: \NRE\ places mass in the correct region but misses the low density region. \SNRE\ fairly accurately reconstructs the posterior, although slightly narrow in some regions.
        \textit{Mid-right column}: \NPE\ produces a wide, poorly resolved posterior. \SNPE\ recovers the low density region but introduces incorrect aberrations.
        \textit{Right column}: \NLE\ fails to accurately place any posterior mass. \SNLE\ accurately reconstructs the posterior.
    }
    \label{fig:other-sbi-torus-corner}
\end{figure}

We applied the various \SBI\ techniques to the torus problem for comparison in Figure \ref{fig:other-sbi-torus-corner}. Just like with \MNRE\ we expected that amortized, non-sequential methods would not have sufficient simulations in the relevant region to accurately model the joint posterior on this task. Similarly, we expected that sequential methods could benefit by focusing samples in the relevant parameter region, thereby learning a more accurate posterior for this piece of data. (Just like \TMNRE.) This is what we observe. Our goal in Section \ref{sec:torus} was to show that truncation offers accuracy with a limited simulation budget, just like sequential methods do.

\subsection{Eggbox details}

The eggbox task is well described in the main text. The hyperparameters for \NRE, \SNRE, and \SMNRE\ are all the defaults as determined by \SBI~\cite{sbi}. We implemented \SMNRE\ by creating a custom version of the simulator. We revealed the one or two parameters which were learned sequentially to the \SMNRE\ algorithm while we ``baked-in'' the uniform prior for the other dimensions, i.e. the simulator sampled from a uniform distribution and simulates a concatenation of the sequentially predicted dimensions with the uniformly predicted dimensions. We expect that \SMNRE\ fails due to its very limited number of simulations. This limitation might seem pathological in this symmetric setting but it is very real in a simulator which defines an unknown posterior that may or may not have symmetry.

\subsection{Alternative Simulation-based Inference Methods on the Eggbox}

\begin{figure}[hbt]
    \centering
    \begin{minipage}{0.24\textwidth}
        \centering
        \includegraphics[width=\textwidth]{eggboxcorner/eggbox-ref-corner.png}
    \end{minipage}
    \begin{minipage}{0.24\textwidth}
        \centering
        \includegraphics[width=\textwidth]{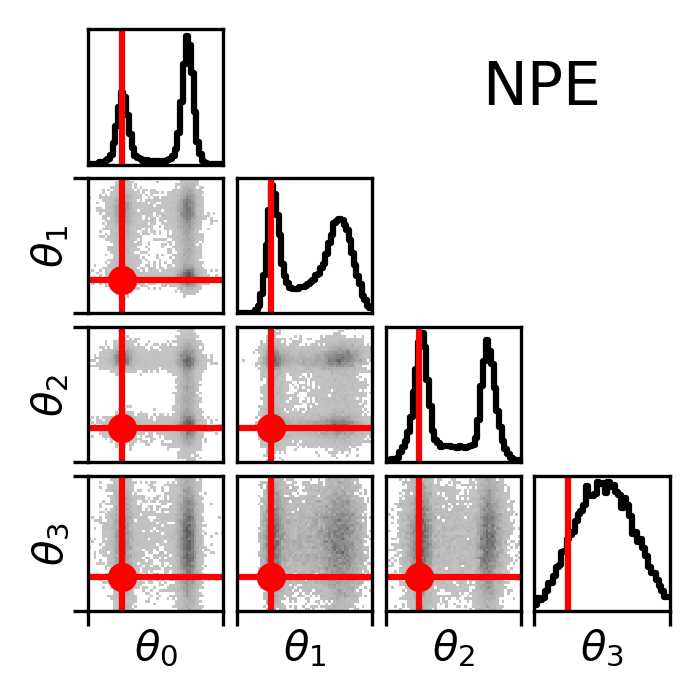}
    \end{minipage}
    \begin{minipage}{0.24\textwidth}
        \centering
        \includegraphics[width=\textwidth]{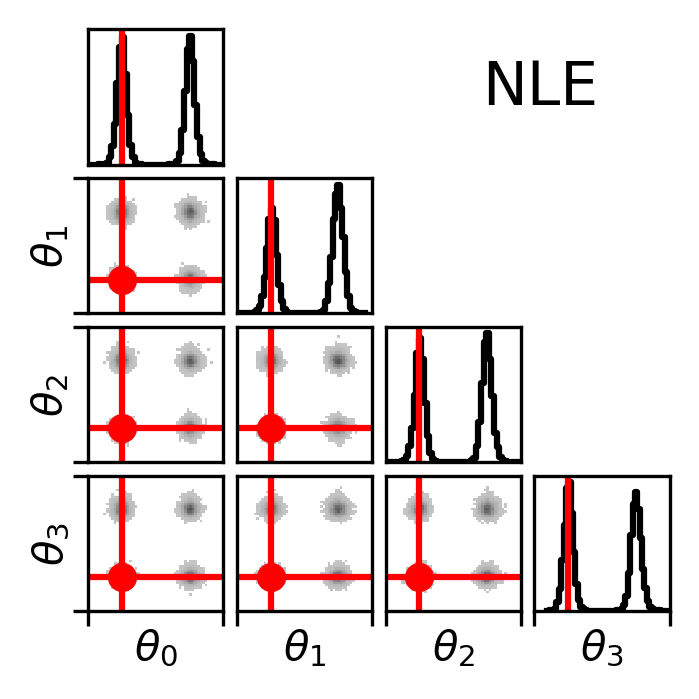}
    \end{minipage}
    \begin{minipage}{0.24\textwidth}
        \centering
        \includegraphics[width=\textwidth]{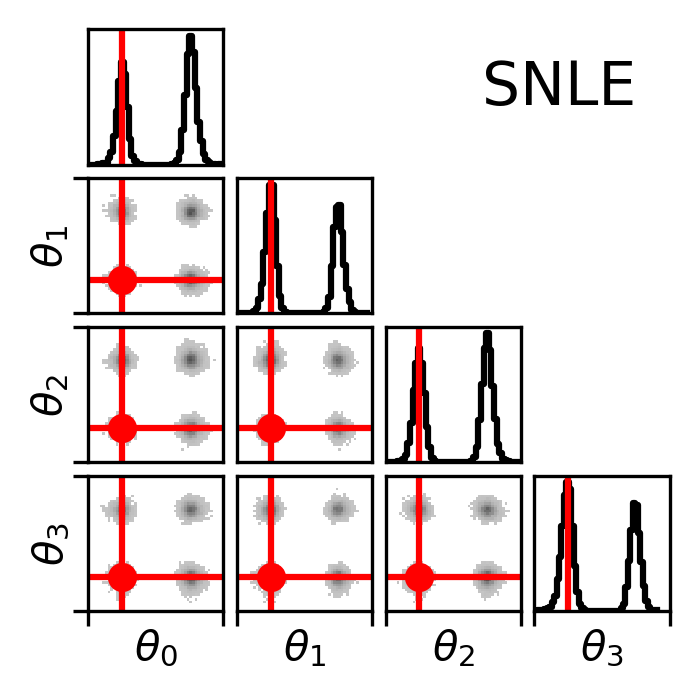}
    \end{minipage}
    \caption{%
        Posteriors from the 10-dim eggbox benchmark with only four parameters shown for clarity. Each method recieved 10k training samples and produced 25k posterior samples. All methods, other than the ground truth, were trained jointly. \NPE\ fails to model the posterior. \SNPE\ training converged, but drawing the posterior samples required well over three days of computation. \NLE\ and \SNLE\ seem to have accurately recovered the ground truth joint distribution with the budget.
    }
    \label{fig:other-sbi-eggbox-corner}
\end{figure}

We applied the various \SBI\ techniques to the eggbox problem for comparison in Figure \ref{fig:other-sbi-eggbox-corner}. The purpose of this task was to show that multimodal distributions can be challenging for methods which estimate the joint. On this problem we found that to be true for \NRE, \SNRE, and \NPE, see figures \ref{fig:other-sbi-eggbox-corner} and \ref{fig:eggbox-corner}. We were unable to draw the necessary samples from \SNPE, likely indicating poor performance.

On the other hand, \NLE\ and \SNLE\ seem to work beautifully for this problem, producing accurate looking posteriors. Likelihood estimation methods generally perform very well on benchmark / toy problems presented in the \SBI\ literature. \NLE\ and \SNLE\ are designed to estimate the data distribution using a flow conditioned on parameters. This works well for low dimensions and simple data; however, flows are notoriously difficult to train on higher-dimensional problems due to the necessary computation of the determinant \cite{papamakarios2019normalizing} which scales with dimension $d$ like $\mathcal{O}(d^3)$. Evidence of this can be seen in the SLCP-Distractors task in Section \ref{sec:sbibm}. Introducing extra dimensionality, even when it contains no useful information for inference, significantly reduced the performance of \NLE\ and \SNLE. Since this work is intended to create a tool for data with $d \gg 100$, we recommend our method over \NLE\ and \SNLE\ until they can be applied to high dimensional problems.

\subsection{Rotated Eggbox}

\begin{wrapfigure}[21]{R}{0.35\textwidth}
	\begin{center}
    	\includegraphics[width=0.35\textwidth]{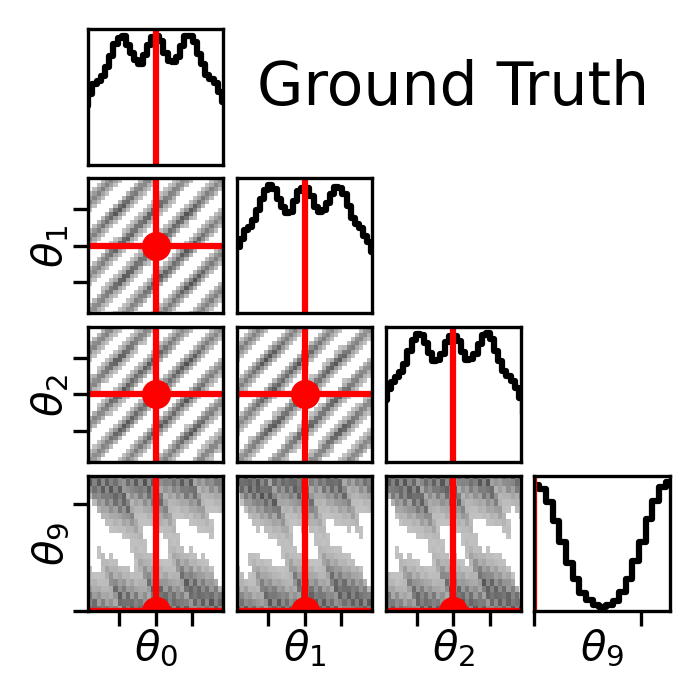}
	\end{center}
    \caption{
        The ground truth of the rotated eggbox posterior estimated by a set of translations and a rotation of samples from the eggbox. The first eight dimensions are symmetric so only two are shown along with the final asymmetric dimension.
	}
	\label{fig:rotated-eggbox}
\end{wrapfigure}

Truncating with one-dimensional marginals may lead to larger volumes than necessary to contain the posterior mass, when it is not axis-aligned. For example, consider the inefficiencies of truncating a highly correlated Gaussian posterior in this way, see Appendix \ref{apndx:limitations}. We wanted to create a posterior which was not axis aligned, thereby simulating a truncation scheme which was forced to truncate a much larger volume than necessary to contain the mass of interest. Our next study considers the eggbox problem transformed by a rotation to remove its axis-alignment.

We created a rotated version of the eggbox simulator where the initial simulator $g(\btheta)$ was replaced by $g(Q^{T} \btheta)$. $Q \in \mathbb{R}^{10 \times 10}$ is a rotation matrix which rotates the point $(1, 1, \ldots, 1)^{T}$ to $(0, 0, \ldots, c)^{T}$ where $c$ is a positive constant. We had been using a uniform prior defined by the unit cube; however, this region no longer holds the posterior mass. We determined the limits of a hyperrectangle which completely covered the rotated prior in the following manner: Consider each unit vector pointing to the edge of the unit cube $E = \times_{n=0}^{10} \{ 0, 1 \}$ where $\times$ denotes the Cartesian product. If each of those vectors $v_i$ is rotated by $Q$, $v_i \mapsto Q v_i$, we can then look at the minimum and maximum values projected along each basis vector and set those to be the limits of our new uniform prior. The bounds were $\left[(-0.924, 0.924), \ldots, (0.000, 3.160) \right]$ which implies a volume 795 times larger than the unrotated prior. We place our new uniform prior over this volume to simulate an inefficient truncation.

Generating samples from the ground truth joint posterior is rather difficult because of the extremely large number of modes and lack of symmetry. To generate samples from the rotated distribution, we simply rotated samples from the original eggbox problem. However, this does not reveal the entire posterior since the simulator is periodic over the prior interval. We first copied the 10,000 samples over all of the hypercubes neighboring the unit cube, yielding 20,470,000 samples, then sub-sampled those and rotated them. More translations were necessary to completely represent the periodic nature of the ground truth in Figure \ref{fig:rotated-eggbox}, but we quickly ran into memory constraints. The limited, single-set-of-neighboring-hypercubes representation of the rotated eggbox is shown in Figure \ref{fig:rotated-eggbox}.

For our experiment we trained all \SBI\ methods along with \MNRE\ on this problem using the same hyperparameters as in the original eggbox. No method was able to faithfully represent the ground truth. We believe that the methods failed due to the extremely large volume of the prior rather than the rotation.

\subsection{Simulation budget and dataset generation}
\label{apndx:data_gen}
First, we note that it is possible to accomplish Algorithm~\ref{alg:tmnre} without having to sample new independently drawn parameters by pairing simulations with other parameters. Like several other algorithms \cite{Hermans2019, Durkan2020}, we assure the independence of $\bx$ and $\btheta$ by sampling two mini-batches from the dataset and switching the $\btheta$ parameters. This produces a pair of independently drawn parameters and simulations which can be used to calculate the loss function efficiently without sampling parameters more than necessary.

Second, we discuss our heuristic for producing a useful amount of samples within the constrained region. We divided the simulation budget between constraining and inference on the constrained region. During the constraining phase, we set the training data per round $N^{(m)} = 0.3 \, B$ where $B$ represents the entire simulation budget. This does not imply that each round used a third of the simulation budget, rather the new simulations plus the retained simulations equal a third of the budget. Finally, once the stopping criterion was satisfied, we used the remaining budget within the estimated $\hat{\Gamma}^{\textrm{rec}}$. We found that this technique created enough simulations during the truncation rounds to estimate $\Gamma^\textrm{rec}$ relatively well while leaving a sizable portion of the simulation budget to be sampled from the truncated prior. Naturally, we want to sample as much from the truncated prior as possible to reduce simulations in regions of nearly zero posterior density and increase simulator efficiency. 

In contrast, sequential methods usually divide their simulation budget evenly across rounds. However, since they do not have a stopping criteria it is natural to divide the simulations that way. We used this technique when training sequential methods.

Third, we discuss the stochastic nature of our sampling technique. Rather than sampling an exact number of parameters and corresponding simulations, we instead sampled from a Poisson distribution centered at the the requested number of samples. In practice, this meant around a 5\% difference between the extrema of the actual number of produced simulations and the requested number of simulations.

\subsection{How do we use the likelihood-to-evidence ratio?}
\paragraph{Histograms.}
Domain scientists, in particularly astronomers and astrophysicists, typically consider a visualization of the posterior and draw conclusions based on their problem-specific intuition and the reported uncertainty bounds. 
By learning the relevant one- and two-dimensional marginal likelihood-to-evidence ratios, our method can generate a visualization of the posterior, namely a corner plot of weighted histograms, by sampling from the prior $\bvartheta \sim p(\bvartheta)$ and sorting the results into bins. 
Each sample's contribution is weighted according to the learned $\rhat(\bx \mid \bvartheta)$, creating a posterior histogram.

The histogram facilitates the computation of credible regions. In particular, finding an accurate estimate of the $(100 - \alpha)\%$ highest density credible region is the primary goal of most astronomers.

\paragraph{Rejection sampling.}
We can use our unnormalized point-wise posterior estimate as the target and the constrained prior as the proposal to generate samples distributed like the posterior via rejection sampling. 
Let 
${\ptilde_{\Gamma}(\bvartheta \mid \bx) = \rhat_{\Gamma}(\bx \mid \bvartheta) \indicator_{\Gamma}(\bvartheta) p(\bvartheta)}$ 
be our target distribution and 
$q(\bvartheta) = \indicator_{\Gamma}(\bvartheta) p(\bvartheta)$ 
be our proposal distribution. 
$\indicator_{\Gamma}$ denotes an indicator function which is nonzero in constrained region $\Gamma$, $\ptilde_{\Gamma}(\bvartheta \mid \bx)$ is the unnormalized posterior, and $\rhat_{\Gamma}$ is the constrained likelihood-to-evidence ratio.
Following a modified version of Maximum Likelihood Estimate (MLE)-based rejection sampling \cite{Smith1992-uc}, we set $M = \rhat_{\Gamma}(\bx \mid \hat{\bvartheta})$ where $\hat{\bvartheta} = \argmax_{\bvartheta} \rhat_{\Gamma}(\bx \mid \bvartheta)$ is the MLE. We sample parameters from the proposal distribution $\bvartheta \sim q(\bvartheta)$ and accept them with probability $\frac{\ptilde_{\Gamma}(\bvartheta \mid \bx)}{M q(\bvartheta)} = \frac{\rhat_{\Gamma}(\bx \mid \bvartheta)}{\rhat_{\Gamma}(\bx \mid \hat{\bvartheta})}$.

The acceptance probability is tolerable when the parameter space is low dimensional and the constrained prior is not significantly wider than the posterior \cite{murphy}.
For our method, the first condition generally holds but the second is not guaranteed.
Despite this potential inefficiency, the parallel proposal and rejection of samples is resolved quickly when the acceptance probability is not vanishingly small.
In that case, likelihood-free \MCMC\ \cite{Hermans2019, sbibm, Durkan2020} becomes unavoidable.

\section{Evaluation metrics}
\label{apndx:evaluation_metrics}

We introduce here the relevant evaluation metrics.  These are relevant both to compare our results with the ground-truth (C2ST, KL divergence), as well as for studying desirable statistical properties of the posterior without requiring knowledge of the ground-truth (self-consistency check / empirical credible interval testing / expected coverage testing).  Here, C2ST is motivated by its omnipresence in the simulation-based inference literature, to which we want to compare. We additionally introduce KL divergence as a metric that is tractable for the low-dimensional marginal posteriors that are the focus of this paper.

The neural likelihood-free inference reports several performance metrics which do not apply well to our method...
Reporting $-\mathbb{E}[\log q(\btheta_{o} \mid \bx_{o})]$ is quite common throughout the literature \cite{papamakarios2019sequential, greenberg2019automatic, Hermans2019, Durkan2020, epsilon_free}. 
Since we learn an unnormalized posterior, we cannot compare our value to other methods. Furthermore it is a poor indicator of performance \cite{sbibm}. 
Another common technique is to measure the median distance between posterior-predictive samples \cite{papamakarios2019sequential, greenberg2019automatic, Durkan2020} but this is impossible since we learn a marginalized posterior and cannot sample from the posterior predictive distribution. 
Maximum Mean Discrepancy (MMD) \cite{gretton2012kernel, ramdas2015decreasing} has been found to be sensitive to choice of hyperparameters \cite{sbibm}.
It is in principle possible to apply other alternatives such at the Wasserstein distance \cite{sriperumbudur2009integral} using the Sinkhorn-Knopp algorithm \cite{cuturi2013sinkhorn} but there is not literature precedent.

\subsection{Kullback–Leibler Divergence}
Since this paper is primarily interested in determining low dimensional marginal posteriors, it is feasible to estimate the Kullback–Leibler divergence, denoted $D_{KL}$, using samples and comparing the histograms. We've found that this method for approximating the Kullback–Leibler divergence is hyperparameter dependent, namely based off the number of bins. We only reported the Kullback–Leibler divergence for the torus problem and we used 100 bins. This effect implies that only the difference between Kullback–Leibler divergences is relevant.

\subsection{Classifier 2-Sample Test per d-Dimensional Marginal (C2ST-ddm)}
The classifier 2-sample test (C2ST) \cite{friedman2003multivariate, lopez2017revisiting} is a performance metric where a classifier is trained to differentiate between samples from the ground truth and approximate posterior. It features an interpretable scale where 1.0 implies that the classifier could distinguish every pair of samples the distributions while 0.5 implies indistinguishably. It is possible to determine where distributions differ using this metric \cite{sbibm}. A classifier with insufficient expressivity yields unreliable results \cite{lopez2017revisiting, sbibm, hermans2021averting}.

We define the \emph{C2ST per $d$-Dimensional Marginal (C2ST-ddm)} test statistic, which reports the average C2ST across every set of $d$-dimensional marginals.
Consider two random variables ${\bX \sim P(\bX)}, {\bY \sim Q(\bY)}$ with $\bX, \bY \in \mathbb{R}^{D}$ and hyperparameter $1 \leq d \leq D$ that represents the marginal dimensionality of interest.
Let
$(S_{P}, S_{Q}) 
\coloneqq 
\left\lbrace 
	(S_{P_k}, S_{Q_k}) : 
	k \in \{1, 2, \ldots, \binom{D}{d}
\right\rbrace$
where ${S_{P_{k}} := \{\bx^{(1)}_{k}, \ldots, \bx^{(n)}_{k}\} \sim P(\bX_{k})}$  and 
${S_{Q_{k}} := \{\by^{(1)}_{k}, \ldots, \by^{(n)}_{k}\} \sim Q(\bY_{k})}$
are sets of $n$ samples drawn from the $k$th $d$-dimensional marginal of $P$ and $Q$, respectively. Now,

\begin{equation}
    \label{eqn:c2st-ddm}
	\text{C2ST-ddm}(S_{P}, S_{Q}) 
	\coloneqq \frac{1}{K} \sum_{k=1}^{K} \text{C2ST} 
	\left(
		S_{P_{k}}, S_{Q_{k}}
	\right),
	\text{ with } K = \binom{D}{d}.
\end{equation}

For our problem, we let $P_{k} = p(\bvartheta_{k} \mid \bx_{o})$ and $Q_{k} = \hat{r}_{k}(\bx_{o} \mid \bvartheta_{k}) p(\bvartheta_{k})$.

\subsection{Empirical Credible Interval Testing}
\label{apndx:ctest}
Evaluating the accuracy of a posterior approximation requires access to the ground-truth and the ability to compute a suitable metric or divergence. While acceptable during benchmarking \cite{sbibm}, this is impossible for practitioners because they only have access to the observation $\bx_{o}$. Domain scientists depend on sanity checks such as coverage testing and comparison between estimation methods to verify that the reported posterior is accurate. Coverage testing is designed for frequentist confidence intervals; however, we apply a similar technique, known in \cite{hermans2021averting} as expected coverage testing, to test the validity of our credible intervals, empirically.

We report a nominal (100 - $\alpha$)\% credible region but the effects of approximation or training might have influenced the contour's shape. 
Our empirical testing checks whether the nominal contour aligns with the contour ground truth by considering many realizations of $\bx$ and dividing the number of times the corresponding $\btheta$ falls within the nominal credible region by the number of $(\btheta, \bx)$s that were tested. When this is the case, the blue line and the orange line intersect in visualizations like Figure~\ref{fig:checks}.

One major advantage of an amortized method for a real-world practitioner is the possibility of quickly performing tests like these.
During the training process many parameter-simulation pairs have already been generated, we can use them to check the credible intervals of our method. 
Note that sequential methods cannot do this without great expense because they would have to retrain their posterior estimator on every tested observation. Furthermore, this test checks the properties of a single amortized estimator; however, testing sequential methods in the same way estimates properties of the sequential training, not a single estimator.

\section{Comparing the truncated marginal likelihood-to-evidence ratio to the truth}
\label{apndx:derivations}

\subsection{Exemplary error estimates}

We will consider the effect of truncation on a multivariate normal distribution, and discuss various limiting cases.  Let us assume that the true posterior has the shape of a multivariate normal distribution with mean zero and covariance matrix $\Sigma$,
\begin{equation}
    p(\bvartheta|\bx_o) = \mathcal{N}(\bvartheta|0, \Sigma)\;.
\end{equation}
This implies that the posterior-to-maximum posterior ratio is given by
\begin{equation}
    \frac{p(\bvartheta|\bx_o)}{\max_{\bvartheta} p(\bvartheta|\bx_o)} =
    \exp\left( -\frac12\bvartheta^T \Sigma^{-1} \bvartheta\right)\;,
\end{equation}
and that the same ratio for all one-dimensional marginal posteriors is given by
\begin{equation}
    \frac{p(\theta_i|\bx_o)}{\max_{\theta_i} p(\theta_i|\bx_o)} =
    \exp\left( -\frac12\frac{\theta_i^2}{\Sigma_{ii}}\right)\;.
    \label{eqn:gauss_example}
\end{equation}
Let us consider an indicator function $\mathbbm{1}_\Gamma$ where $\Gamma$ is defined as in Eq.~\eqref{eqn:GammaRec}, given some small $\epsilon$.  In the case of vanishing parameter correlations, the covariance matrix $\Sigma$ is diagonal, and the posterior factorizes like $p(\bvartheta|\bx_o) = \prod_i p(\theta_i|\bx_o)$.  The truncation procedure can then be considered for each of the one-dimensional marginal posteriors separately:  Only parameter regions where $|\theta_i| < \sqrt{-2 \Sigma_{ii} \ln \epsilon}$ for all $i$ are included in $\Sigma$.  In this case $\epsilon$ directly determines how far into the tails posteriors are correctly reconstructed.   Using the error function, one can show that the amount of mass that is removed by the truncation is $\epsilon/\sqrt{-\ln\epsilon}$.  This motivates our general estimate of an $\mathcal{O}(\epsilon) \max_{\theta_i} p(\theta_i|\bx_o)$ effect on the truncated posteriors, where the second factor is accounting for the right dimensionality of the expression.

Let us consider the opposite extreme of a maximally correlated posterior, with a covariance matrix that is given by $\Sigma_{ii} = 1$ and $\Sigma_{ij} = 1 - \xi$ for $i\neq j$, and where $\xi \ll1$.  Again, marginal posteriors are given by Eq.~\eqref{eqn:gauss_example}.  Since the support of the maximally correlated posterior is essentially focused on the line $\theta_1 \simeq \theta_2 \simeq \dots \simeq \theta_d$, truncations in all directions are identical.  As a result, marginal posteriors are affected exactly as in the previous diagonal case.

Finally, let us consider a mildly correlated posterior in two dimensions. In this case, the region $\Gamma$ would be again identified through $|\theta_i| < \sqrt{-2 \Sigma_{ii} \ln \epsilon}$ for $i=1,2$, but since the posterior does not factorize anymore integrals on the constrained region become non-trivial.  However, since only $\mathcal{O}(\epsilon)$ of posterior mass lies outside of $\Gamma$, this implies that only a similarly small mass fraction can be re-distributed in the truncated marginal posteriors $p_\Gamma(\theta_i| \bx_o)$.  This can significantly affect the far low-mass tails of the distribution, with negligible effect on the high mass density regions of the posterior.

\subsection{A general estimate}

Let us consider an indicator function defined through Eq.~\eqref{eqn:GammaDef}, first for a single marginal $\theta_i$.  The removed probability mass is then given by 
$$
\delta M_\epsilon = \int_{\Omega_i} d\theta_i p(\theta_i|\bx_o) \indicator \left[p(\theta_i|\bx_o) < \epsilon \max_{\theta_i} p(\theta_i|\bx_o)\right]\;,
$$
where $\indicator$ denotes an indicator function.  An upper bound on the removed probability mass is then given by 
$$
\delta M_\epsilon < \epsilon \max_{\theta_i} p(\theta_i|\bx_o)\int_{\Omega_i} d\theta_i  \indicator\left[p(\theta_i|\bx_o) < \epsilon \max_{\theta_i} p(\theta_i|\bx_o)\right]\;,
$$
For a compact $\Omega_i$, this leads to the claimed bound in one dimension.  However, also in the case of a larger number of parameters, each truncation would remove at most mass at the level of $\mathcal{O}(\epsilon)$, leading to an overall $\mathcal{O}(\epsilon)$ effect on the estimated posteriors.
We emphasize that in the case of priors with non-compact support, a re-parametrization onto priors with compact support can lead to smaller coefficients in front of $\epsilon$.

\section{Limitations}
\label{apndx:limitations}

We note two kinds of limitations: First, we address limitations when the method works as planned. Second, we address failure modes.

When the posterior distribution is nearly as wide as the prior, we do not gain much by truncating the prior distribution. In this case our method would reduce to \MNRE. However, this is rarely the case in physics where the paradigm is to define an uninformative prior distribution across the accepted bounds for a parameter and the posterior will be contained in fractions of that prior's mass. 

Another limitation is that the truncation by hyperrectangle is inherently inefficient when the marginals of interest are highly correlated. In that situation, we are interested in a hyperellipse within the constrained hyperrectangle but our current formulation cannot utilize this heuristic. This problem possible to solve by using techniques from Nested Sampling which regularly seeks to efficiently sample from within a certain density contour.

Finally, as mentioned in Section~\ref{sec:tmnre}. Algorithm~\ref{alg:tmnre} does not naively allow for sampling from the posterior predictive distribution $p(\bx' \mid \bx) \coloneqq \int_\Omega p(\bx' \mid \btheta) p(\btheta \mid \bx) \, d\btheta$ because it only produces the one- and two-dimensional marginals. It remains possible to estimate the joint posterior within the truncated region by simply training another ratio estimator with all parameters. Doing this, using only the simulations necessary for Algorithm~\ref{alg:tmnre}, may produce an inaccurate joint posterior estimate.

The failure modes are perhaps more obvious. If our initial round of sampling is too sparse, it is possible to incorrectly ``miss'' a region of high posterior density and cut it out of our analysis. If the initial region is satisfactorily sampled from, this will not occur. To illustrate this point, consider a simulator with a two-dimensional parameter space. If significant amounts of posterior mass are truncated in the $\theta_0$ dimension, the ground truth $\theta_1$-marginal posterior, under the truncated prior, transforms from the intended marginal distribution $p(\theta_1 \mid \bx)$ towards a conditional distribution $p(\theta_1 \mid \bx, \theta_0^\ast)$. Where $\theta_0^\ast$ denotes the center of the truncated region in the $\theta_0$ dimension.

Another failure mode is related to the local amortization that our ratio estimators learn. While they are able to estimate any posterior from a parameter drawn within the truncated prior, it may be that some of the posterior runs into the truncation. This can be identified whenever a posterior equicontour line intersects with the truncation bounds. The caveat is that if an entire separate mode is truncated, this test will not indicate it. In general, we suggest limiting the use of the locally amortized predictions to ones closer to the ground truth generating parameter than to the truncation bounds.

\section{Cosmological inference with a simulator}
\label{apndx:cosmology}
\begin{figure}[ht]
    \centering
    \includegraphics[width=\textwidth]{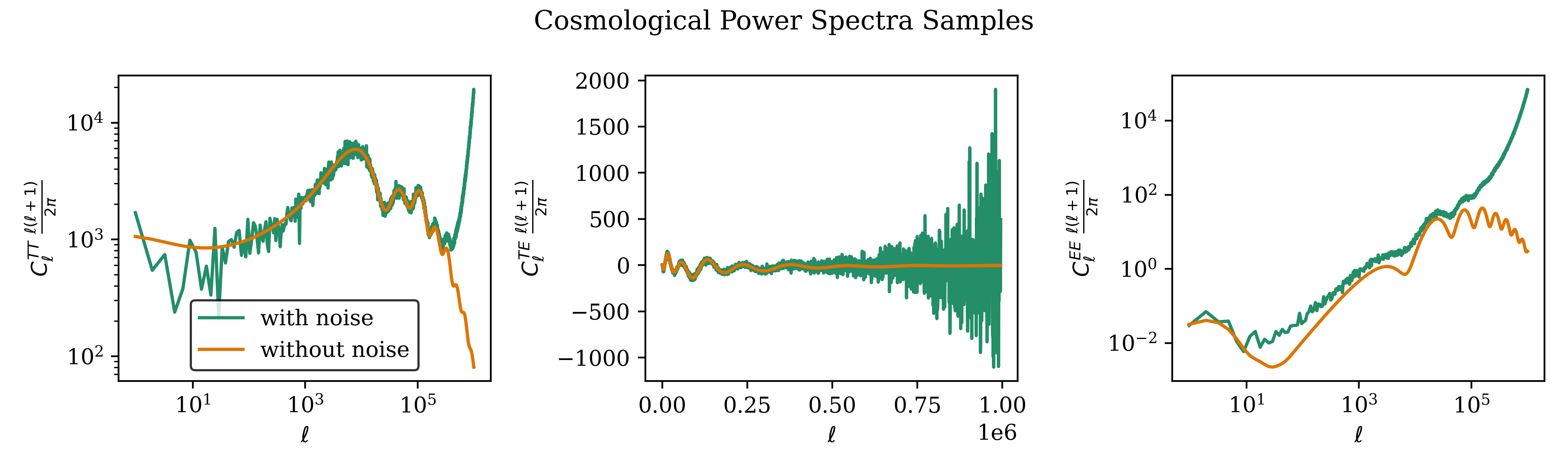}
    \caption{A sample drawn from the CMB simulator, with the cosmological contribution in orange and the noise-added sample in green. The noise model amounts to a non-stochastic contribution from the instrument as well as a stochastic contribution (following a Wishart distribution) corresponding to cosmic variance, in other words the fact that an individual $C_\ell$ is determined by measuring $2\ell+1$ modes.}
    \label{fig:TTTEEE_sample}
\end{figure}

Parameter inference plays a important role in modern cosmology. Here we use a simulator that takes six parameters (specifying the underlying $\Lambda$CDM cosmological model) and returns three lensed angular power spectra $C_{\ell}^{TT},C_{\ell}^{TE},C_{\ell}^{EE}$ (where $T$ denotes temperature and $E=$ denotes E-mode polarization) as they would be measured by an idealized Cosmic Microwave Background (CMB) experiment. The likelihood-based approach to inference in this context is provided by popular packages such as \texttt{MontePython} \cite{Brinckmann:2018cvx}. Our simulator is identical to the likelihood that in \texttt{MontePython} is called \texttt{fake\_planck\_realistic} \cite{DiValentino:2016foa}. This likelihood is used, often in combination with other likelihoods, to forecast the expected constraining power of future experiments. In this model, the power spectra receive non-stochastic contributions from the cosmological model and the idealized measurement instrument. Stochasticity is implemented in the form of \emph{cosmic variance}, which reflects the fact that for fixed $\ell$, each $C_{\ell}$ is determined by measuring $2\ell+1$ modes in the sky. The result is that the collection of $C_\ell$ obeys a Wishart distribution, which at large $\ell$ can be approximated as a multivariate normal distribution. For more details see \cite{Perotto:2006rj}. Draws from the simulator with and without noise are shown in Fig.\ \ref{fig:TTTEEE_sample}.

We use this example to study the utility of \emph{marginal} ratio estimation. As such, we do not use multiple rounds of simulation and training. This is in part due to the availability of a tractable likelihood, which allows us to perform a Fisher estimation (i.e.\ Gaussian approximation) of the expected marginal probability contours. Although the ground-truth posteriors for this inference problem turn out to be slightly non-Gaussian, the Fisher estimation suffices to derive a reasonable region in parameter space for inference. We therefore take a uniform prior with $\theta_{d}\in \left[\overline{\theta}_{d}-5\sigma^F_{d},\overline{\theta}_{d}+5\sigma^F_{d}\right]$, where $\overline{\theta}_{d}$ denotes the ground truth parameter value and $\sigma^F_{d}=\sqrt{(F^{-1})_{ii}}$ is the Fisher estimation of the 1$\sigma$ region for parameter $i$.

We compare three approaches with 5,000 samples. For comparison, an \MCMC\ analysis of this problem converges after roughly 45,000 accepted samples with an acceptance rate of $\sim 0.3$. We compare \MNRE, \NRE, and \MCMC\ with a limited number of samples. For \MCMC\, we use a pre-computed covariance matrix for proposal steps, determined by running a chain until convergence. For inference with \MNRE~ and \NRE, we use a linear compression layer that takes the concatenated power spectra (each with $\ell \in [2,2500]$, so that the full data vector has 7497 entries) and outputs 10 features. The same linear compression network is shared between different ratio estimators. In other words, we introduce a shared feature embedding of the data such that the entire neural network has the form

\begin{equation}
	f_{\phi, k}(\bx, \bvartheta_{k}) = g_{\phi_{g}, k}(\boldsymbol{F}_{\phi_{\boldsymbol{F}}}(\bx), \bvartheta_{k})
\end{equation}

where $\boldsymbol{F}$ is the feature embedding, $k$ represents the index of the marginal-of-interest, $g$ is an MLP, and $\phi$ represents the network weights from both $g$ and $f$. This is appealing computationally but, unlike with $\phi_{g}$, the weights of the feature embedding are dependent on the loss of every marginal. This is the multi-target training paradigm and can be difficult to tune \cite{degrave_korshunova_2021, Platt1988-dc}; however, this is not a problem for us in practice. The hyperparameters are written in Table~\ref{table:physics_hyperparameters}.

\begin{figure}[ht]
    \centering
    \includegraphics[width=0.33\textwidth]{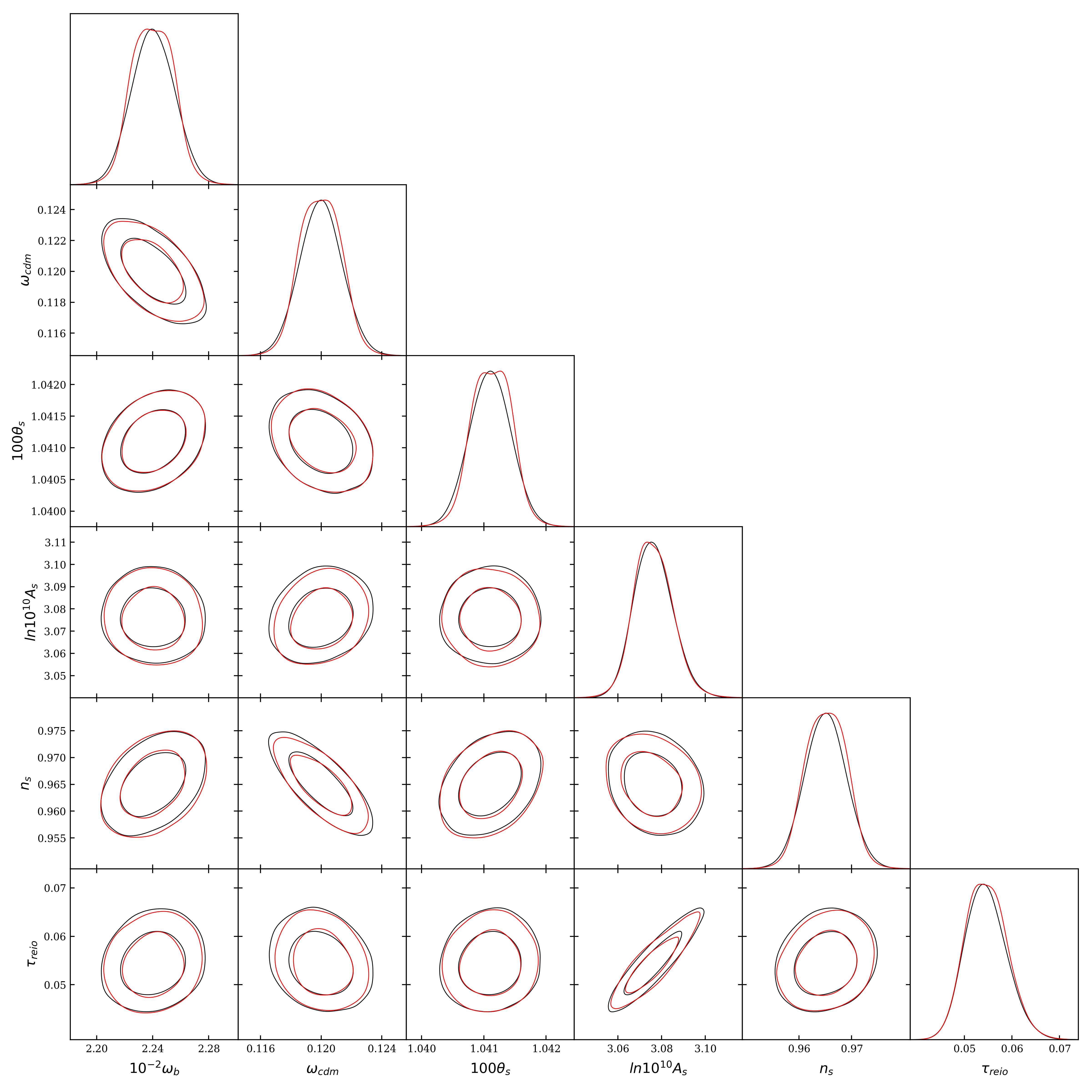}
    \includegraphics[width=0.33\textwidth]{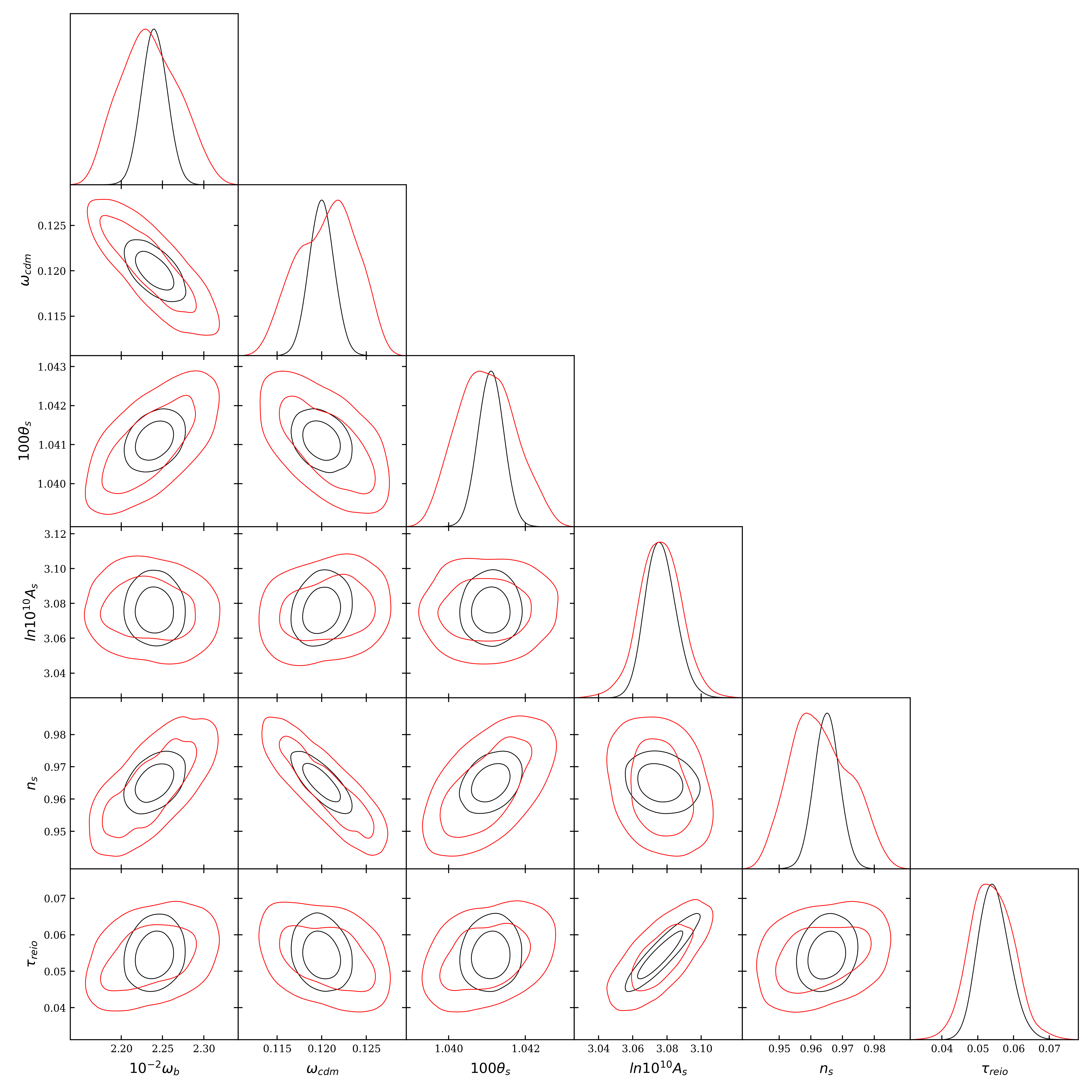}
    \includegraphics[width=0.33\textwidth]{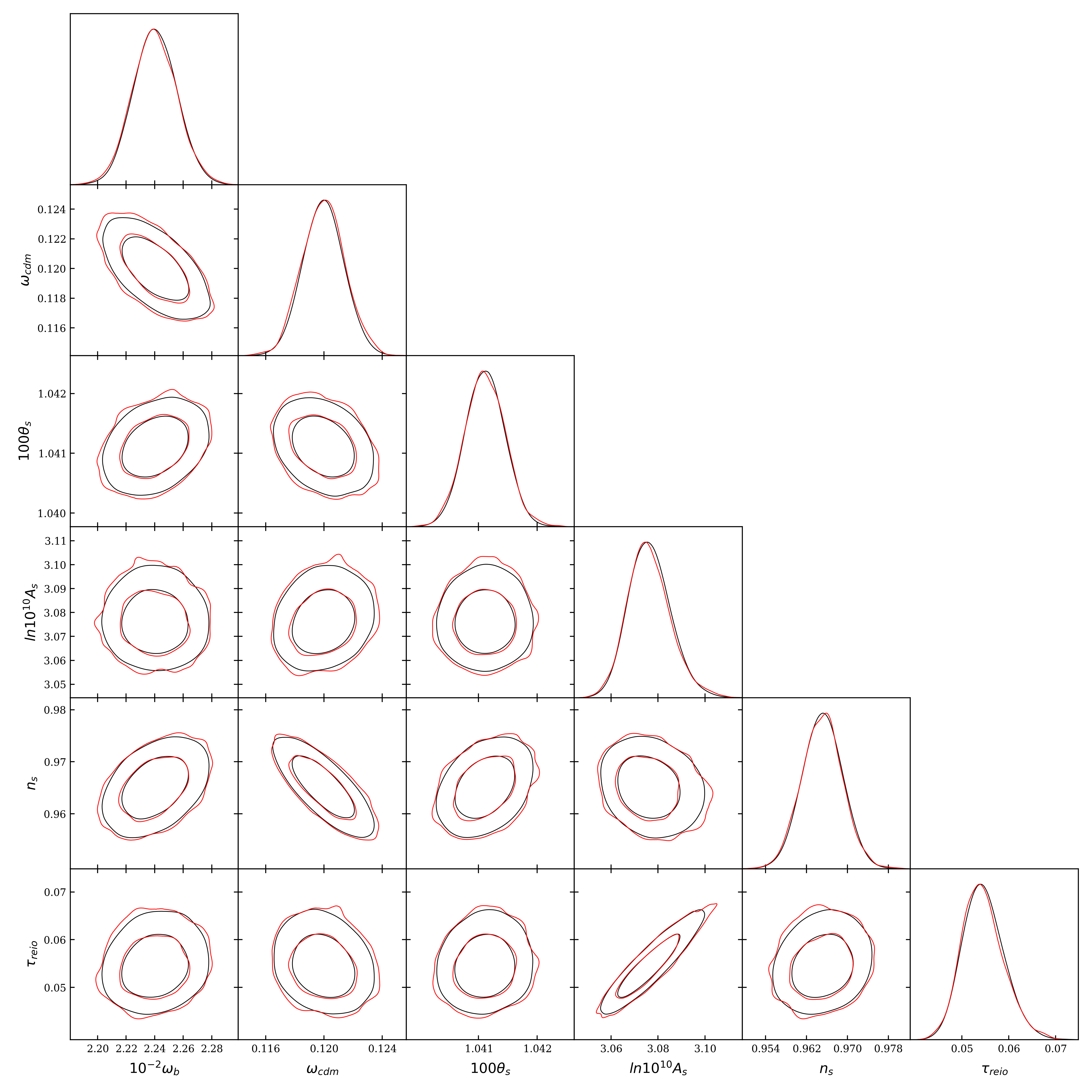}
    \caption{
        Corner plots for various methods using 5,000 simulations (in red) vs.\ ground-truth \MCMC\ (45,000 accepted samples with acceptance rate $\sim 0.3$, in black). Top Left: results for \MNRE. We see excellent agreement with the ground truth. Top Right: corner plot for \NRE. With 5,000 samples, the marginal posteriors are hardly constrained. Bottom: corner plot for \MCMC\ with 5,000 accepted (burn-in removed) samples vs.\ converged \MCMC\ chain. While the short chain gives accurate 1$\sigma$ contours, it does not yield accurate for the 2$\sigma$ contours.
    }
    \label{fig:TTTEEE_corner}
\end{figure}

The results are shown in Fig.\ \ref{fig:TTTEEE_corner}. We see that \MNRE\ reproduces the ground-truth 1- and 2-$\sigma$ contours very accurately. On the other hand, \NRE\ results in hardly any constraint on the parameter space, while the limited \MCMC\ run does not have accurate 2-$\sigma$ contours.

\begin{table}[htb]
	\caption{Physics Example Hyperparameters}
	\label{table:physics_hyperparameters}
	\centering
    \begin{tabular}{ll}
        \toprule
        Hyperparameter & Value \\
        \midrule
        Activation Function & Feature Embedding: None, Ratio Estimator: \textsc{relu} \\
        \textsc{amsgrad} & No \\
        Architecture & Feature Embedding: One Linear Layer, Ratio Estimator: MLP \\
        Batch size & 64 \\
        Batch normalization & No \\
        Criterion & \textsc{bce} \\
        Dropout & No \\
        Early stopping patience & 5 \\
        $\epsilon$ & N/A  \\
        Hidden features & 256 \\
        Percent validation & 10\% \\
        Reduce lr factor & 0.25 \\
        Reduce lr patience & 2 \\
        Max epochs & 300 \\
        Max rounds & N/A \\
        Learning rate & 0.001 \\
        Learning rate scheduling & Decay on plateau \\
        Optimizer & \textsc{adam} \\
        Weight Decay & 0.0 \\
        Z-score observations & online \\
        Z-score parameters & online \\
        \bottomrule
    \end{tabular}
\end{table}

\end{document}